%% file: arxiv.tex
\documentclass[runningheads]{llncs}

\usepackage[mobile]{eccv}

\usepackage{eccvabbrv}

\usepackage{graphicx}
\usepackage{booktabs}

\usepackage[accsupp]{axessibility}  % Improves PDF readability for those with disabilities.

\usepackage{hyperref}

\usepackage{makecell}
\usepackage{float}
\usepackage[numbers]{natbib}
\usepackage{enumitem}
\usepackage[table]{xcolor}
\usepackage{setspace}
\usepackage{amsmath}
\usepackage{amssymb}
\newcommand{\customfootnotetext}[2]{{% Group to localize change to footnote
\renewcommand{\thefootnote}{#1}% Update footnote counter representation
\footnotetext[0]{#2}}}% Print footnote text

\usepackage{orcidlink}
\usepackage{pifont}
\usepackage{fontawesome5}    
\usepackage[most]{tcolorbox} 
\usepackage{todonotes}
\usepackage{multirow}
\usepackage{wrapfig}
\usepackage{hyphenat}

\definecolor{MyDarkGreen}{RGB}{204,85,0}

\newtcbox{\badge}{on line,
  colback=blue!15, colframe=blue!40!black,
  boxsep=1pt, left=2pt, right=2pt, top=0.5pt, bottom=0.5pt,
  arc=2pt,                    
}

\begin{document}

\title{UniCanvas: A Diffusion-based Unified Model \\ for Text-in-Image Joint Generation} 

\titlerunning{UniCanvas: Diffusion-base Unified Model for Text-in-Image Joint Generation}

\author{Zeyuan Yang\inst{1}$^*$ \and Hao-Wei Chen\inst{1}$^*$ \and Xueyang Yu\inst{1} \and Yuncong Yang\inst{1} \and Haoyu Zhen\inst{1} \and Ziqiao Ma\inst{2} \and Maohao Shen\inst{3} \and Chuang Gan\inst{1}}

% % TODO FINAL: Replace with an abbreviated list of authors.
\authorrunning{Z.~Yang et al.}

% % TODO FINAL: Replace with your institution list.
\institute{$^1$UMass Amherst \quad $^2$University of Michigan \quad $^3$MIT}

\maketitle
\customfootnotetext{*}{Equal contribution.}
% \customfootnotetext{$\dag$}{This work was done when two of the authors were remote interns at UMass.}

\input{sections/0_abstract}
\input{sections/1_intro}
\input{sections/2_background}
\input{sections/3_method}
\input{sections/4_experiment}
\input{sections/5_conclusion}

\section{Acknowledgement}

We are extremely grateful to Jiaben Chen, Bairu Hou, Sunli Chen, Lixing Fang, and Ziwei Liu for their helpful feedback and insightful
discussions.

\clearpage
\input{sections/a_supplement}

\clearpage
\bibliographystyle{splncs04}
\bibliography{main}

\end{document}

%% file: sections/0_abstract.tex
\begin{abstract}
Recent years have seen remarkable progress in unified vision-language models handling both multimodal understanding and generation within a single architecture. While autoregressive VLMs can reason across modalities, they fail to generate high-quality images. In contrast, diffusion models produce photorealistic visuals yet struggle to generate coherent text, making it challenging to develop a single unified model that can seamlessly handle both visual and text generation. Recent advances suggest that language can be effectively embedded within visual representations, allowing models to reason about textual semantics directly from images. To this end, we propose \textbf{\model}, a first attempt that unifies diffusion models to generate interleaved multimodal contents through text-in-image generation. Diffusion models naturally capture transformations on a shared pixel canvas, which can be viewed as world models of visual change. Instead of producing discrete text tokens, the model learns to represent language as visual patterns inside images, leveraging its inherent multimodal embedding space. This design allows the model to ``draw'' text naturally within a single pixel canvas during image synthesis, achieving seamless multimodal generation. Experiments demonstrate that \model improves performance over previous unified models, positioning text-in-image generation with diffusion models as a promising unified multimodal generation paradigm.
\keywords{Unified Multimodal Model \and Image Editing \and Multimodal Generation}
\end{abstract}

%% file: sections/1_intro.tex
\section{Introduction}
\label{sec:intro}
\begin{figure*}[!t]
    \centering
    % \footnotesize
    \scriptsize
    \includegraphics[width=0.95\linewidth]{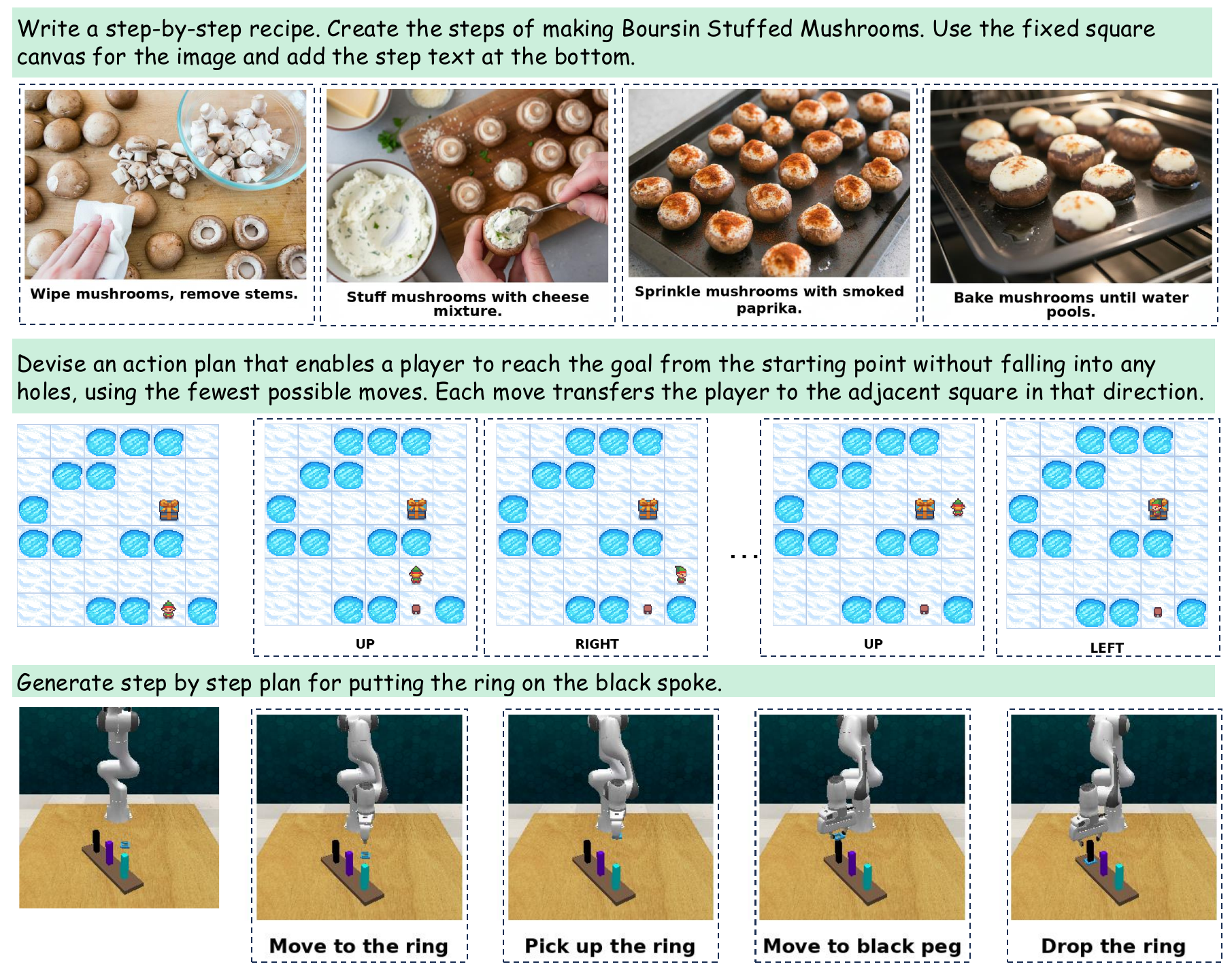}
    \caption{\textbf{Sequential multimodal reasoning across diverse tasks.} Sequential evaluation results of our model across three datasets: Recipe, FrozenLake, and RLBench. The green-shaded regions show the context prompts provided to the model, while the text beneath each image represents optical text inferred through visual reasoning. Our model generates coherent interleaved image–text sequences for diverse tasks, including cooking-step generation, action planning in navigation, and robotic manipulation.
    % Examples of \textbf{\model} generating coherent interleaved image and text sequences.
    }
    \label{fig:teaser}
\end{figure*}

Recent advances in diffusion models and vision–language models (VLMs) have transformed multimodal learning. Autoregressive VLMs excel at cross-modal reasoning and linguistic understanding, yet struggle to synthesize high-quality, spatially faithful images. Diffusion models, in contrast, model complex image distributions and produce photorealistic content with strong spatial coherence, but offer limited step-by-step reasoning. Unifying these complementary strengths in a single architecture remains an open challenge.

Most existing unified models still treat text and image generation as separate end tasks. Text is produced as discrete tokens with an autoregressive head, while a distinct visual decoder synthesizes images. This split induces three gaps: (i) an objective gap—text is learned with a next-token prediction loss whereas images use reconstruction or denoising objectives; (ii) a representation gap—text lives in a symbolic token space while images live in pixels or visual tokens; and (iii) an inference gap—reasoning unfolds primarily in the language head, largely decoupled from the image generator.

Imagine instead a model endowed with a shared, writable canvas, a space where both language and vision can be expressed and refined together. Rather than producing text and images through separate heads, the model can ``write'' and ``draw'' directly on the same pixel-native canvas, which allows linguistic and visual elements to influence each other at every generation step. In this framework, text is no longer treated as a sequence of discrete tokens but as pixel-rendered symbols generated by the same mechanism that synthesizes images. This unified pixel-native representation enables seamless interaction between reasoning and rendering, where words and visuals co-evolve as parts of a single multimodal output.

Guided by this, we introduce \textbf{\model,} a single-backbone diffusion model that represents both text and images on a shared pixel canvas and learns them jointly under a unified denoising objective. Unlike prior multimodal systems that separate text decoding from image synthesis, \model performs \textit{text-in-image joint generation}. It renders legible text directly within images, preserving the spatial structure and photorealistic quality characteristic of diffusion models. During generation, the model progressively refines the canvas across multiple diffusion steps. Each intermediate canvas encodes the current state of a multi-step task, and the model is trained to map from the current canvas and instruction to the next one to update, overwrite, and extend it as the task unfolds.
Diffusion models naturally capture visual transformations on a shared pixel canvas, serving as world models for visual change and enabling step-by-step state updates during generation.
In this way, the multimodal canvas acts as an externalized trajectory of task states, enabling \model to keep track of task progress and thereby support multi-step reasoning within the same generative process.

We validate this on multi-step visual tasks such as VSP, where the model must iteratively update a scene according to a sequence of instructions. \model consistently outperforms strong unified baselines that rely on token-based text heads and separate image decoders, achieving higher accuracy and better temporal consistency across steps. Qualitatively, the evolving canvases produced by \model exhibit coherent state transitions, both in the spatial layout of objects and in on-canvas textual annotations, highlighting the benefits of reasoning directly in a shared pixel space.
Our main contributions are three-fold:
\begin{itemize}
\item We propose a purely diffusion-based multimodal generation framework that operates entirely on a shared pixel canvas, enabling interleaved image and text generation without the need for external VLMs, autoregressive token decoding, or discrete text representations.
\item We introduce a two-stage generation paradigm that separates reasoning from visual synthesis while sharing a single backbone, and a lightweight CLIP-based alignment loss that improves in-image text readability and consistency.
\item We demonstrate that \model produces more readable and consistent text and achieves substantially better performance on multi-step benchmarks compared to existing unified baselines, while also providing initial evidence that \model can support nontrivial textual reasoning beyond fixed action spaces, suggesting a promising direction for further scaling reasoning.
\end{itemize}

\begin{table*}[!t]
    \centering
    \caption{\textbf{Comparison with previous unified models.} \textbf{w/o VLM} indicates whether the model relies on a VLM to generate text. \textbf{Native} specifies if the model is a single architecture. \textbf{w/o Token-based} denotes whether the model relies on a token-based modeling approach. \textbf{Text} specifies if the model can generate explicit text. \textbf{Interleaved} represents whether the model supports multimodal generation.}
    \scriptsize
    \begin{tabular}{@{}lccccc@{}}
        \toprule
        Models & \textbf{w/o VLM} & \textbf{Native} & \textbf{w/o Token-based} & \textbf{Text} & \textbf{Interleaved} \\
        \midrule
        MM-Interleaved~\cite{tian2024mm} & \xmark & \xmark & \xmark & \cmark & \cmark \\
        DreamLLM~\cite{dong2023dreamllm} & \xmark & \xmark & \xmark & \cmark & \cmark \\
        Anole~\cite{chern2024anole} & \xmark & \cmark & \xmark & \cmark & \cmark \\
        MMaDA~\cite{yang2025mmada} & \cmark & \cmark & \xmark & \cmark & \xmark \\
        VPRL~\cite{xu2025visual} & \xmark & \cmark & \xmark & \xmark & \xmark \\
        BAGEL~\cite{deng2025emerging} & \xmark & \cmark & \xmark & \cmark & \cmark \\
        \midrule
        \model & \cmark & \cmark & \cmark & \cmark & \cmark \\
        \bottomrule
    \end{tabular}
    \label{tab:baseline-feature-comparison}
\end{table*}

%% file: sections/2_background.tex
\section{Related Work}
\label{sec:background}

\subsection{Diffusion Models}
\label{subsec:diffusion}
Diffusion models~\cite{sohl2015deep, ho2020ddpm, song2020score} have recently become a dominant paradigm for data generation, which operates by iteratively refining random noise into meaningful structure through a learned reverse diffusion process. They now represent the state of the art in both image synthesis~\cite{dhariwal2021diffusion, karras2022elucidating, yin2025reasonedit, wang2026promptrl} and probabilistic density modeling~\cite{kingma2021variational, nichol2021improved, song2021maximum}. Building on this foundation, conditional diffusion models~\cite{ho2022classifier, chen2026show} have proven especially effective for cross-modal generation, which serves as a powerful framework for translating between textual and visual domains. These models~\cite{chen2023pixart, gao2024lumina, podell2023sdxl, ramesh2022hierarchical, rombach2022high, saharia2022photorealistic} produce highly realistic images from textual descriptions, showcasing impressive alignment between semantics and visual fidelity.

\subsection{World Models }
\label{subsec:world_models}
World Models~\cite{ha2018world} have emerged as widely adopted frameworks in control~\cite{qian2025wristworld}, planning~\cite{yang2025mindjourney} and navigation~\cite{bar2025navigation} for learning predictive transition functions. Most prior approaches parameterize dynamics in compact latent spaces~\cite{hafner2019dream, gao2025adaworld, zhen2025tesseract, zhen20243d} and are primarily developed for policy optimization for embodied intelligence~\cite{lingbot-va2026, ye2026worldactionmodelszeroshot, liao2025genie}. More broadly, a world model can be understood as a learned operator that updates a state representation under structured interventions~\cite{zhang2026foreact, teoh2025next} and such interventions could correspond to visual transformations~\cite{zeng2025editworld}. From this perspective, a model that can apply transformation-conditioned updates to a visual state effectively performs state simulation in pixel space. Our approach adopts this viewpoint and treats visual generation as an explicit state-transition process, enabling step-by-step reasoning, analogous to action-conditioned rollouts in classical world models.

\subsection{Unified Multimodal Models}
\label{subsec:multimodal}
Recent advances in large multimodal~\cite{bai2025qwen25vl} and visual generative models~\cite{esser2024scaling} have motivated frameworks that jointly support perception and generation within a shared architecture~\cite{chen2025blip3, tong2024metamorph, wu2024vila}. Existing works mainly fall into three categories, based on their design principles: autoregressive models~\cite{chern2024anole, team2024chameleon}, diffusion arch~\cite{yang2025mmada, wang2025fudoki, shi2025muddit} and hybrid system~\cite{xie2025showo, zhou2025transfusion, ma2024janusflow}. Despite the progress in architecture evolvement~\cite{ma2025unitok, deng2025emerging} and pre-train / post-train techniques~\cite{jiang2025co, qin2025unicot}, understanding and generation are still largely handled through distinct encoder–decoder pathways. Whether these two capabilities can truly co-exist and mutually enhance each other remains an open question. Few recent works~\cite{guo2025can,guo2025thinking, jiang2025draco, han2026unicorn}, explore how improved understanding benefits generation~\cite{duan2025got, xiao2025mindomni, jiang2025t2i, su2026generation}. However, comparatively little attention has been paid to the reverse direction, how generation may facilitate understanding~\cite{gu2025thinkmorph, wu2026visual, yang2025machine}. A substantial gap remains toward interleaved multimodal generation that tightly integrates perception and synthesis.
In this work, we investigate a new direction towards this gap: unifying understanding and generation within a shared space, and text reasoning is performed directly in pixel space via text-in-image modeling.

%% file: sections/3_method.tex
\section{Method}
\label{sec:method}

\begin{figure*}[t]
\centering
\scriptsize
\includegraphics[width=0.95\linewidth]{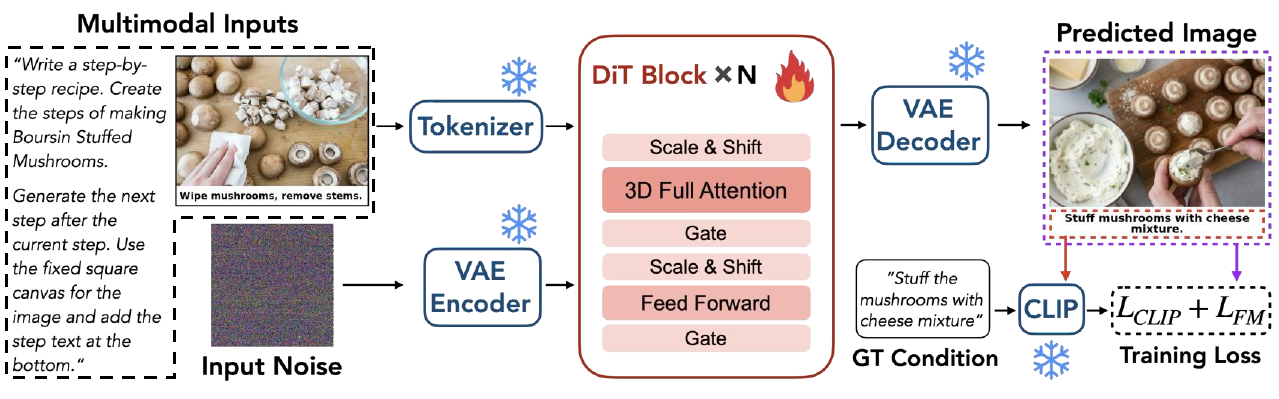}
\caption{\textbf{Overall Architecture of \model.} Our model encodes multimodal inputs and input noise through a frozen tokenizer and VAE encoder, then processes them with stacked DiT blocks. A frozen VAE decoder reconstructs the predicted image, and CLIP-based losses enforce alignment with the ground-truth textual condition.}
\label{fig:ted_arch}
\end{figure*}

At the core of \model is a shared pixel-native canvas: a writable latent space where language and vision share a unified representation and are updated jointly. Instead of using a text head plus a separate image decoder, \model treats language understanding and generation as visual composition on this canvas. A single diffusion backbone learns to: (i) “write’’ text as structured optical patterns encoding semantics, (ii) update the canvas into the next scene state for reasoning. Because all updates operate on the same canvas, the model can be applied repeatedly for multi-step tasks, with each snapshot capturing the evolving state.

In this section, we first describe how \model realizes text-in-image generation via flow matching on the pixel canvas (Sec.~\ref{subsec:text-in-image}). We then explain the two-stage procedure that separates visual reasoning and visual synthesis while sharing a single diffusion backbone (Sec.~\ref{subsec:two-stage}). Next, we introduce a lightweight CLIP-based regularization term that improves text-image coherence and readability (Sec.~\ref{subsec:regularization}). Finally, we briefly describe the pretraining stage that endows \model with foundational text rendering capability on the canvas (Sec.~\ref{subsec:pretrain}).

\subsection{Text-in-Image Generation}
\label{subsec:text-in-image}

In text-in-image generation, \model learns to ``write'' linguistic content directly onto the shared pixel-native canvas. Given an input image $x$, we define a target latent $x^{\text{text}}_0$ in which textual concepts have been visually materialized as optical words and phrases (\eg, action labels, ingredients, instructions, or object relations) rendered on the canvas as part of the scene. These structured visual patterns encode linguistic meaning while remaining compatible with surrounding appearance like layout, color, and local texture.

We adopt a flow-matching formulation to learn the transformation from $x$ to $x^{\text{text}}_0$. Given source and target latents $x_0$ and $x_1$, we sample $t \sim \mathcal{U}[0,1]$ and construct a linear interpolation $x_t = (1-t)x_0 + t x_1$,
and train a velocity network $v_{\theta}(x_t, t, c)$ to predict the true velocity $v_t = x_1 - x_0$. For text-in-image generation we set $x_0 = x$, $x_1 = x^{\text{text}}_0$ and use a reasoning-oriented context prompt $c_{\text{reason}}$, yielding the flow-matching loss
\begin{equation}
\mathcal{L}^{\text{text}}_{\text{FM}} =
\mathbb{E}_{t, x, x^{\text{text}}_0}
\left[
\lVert v_{\theta}(x_t, t, c_{\text{reason}}) - (x^{\text{text}}_0 - x) \rVert^2
\right].
\label{eq:flow_matching_text}
\end{equation}

At inference time, we start from the input image and numerically integrate the learned velocity field backward from $t=1$ to $t=0$ for $T_{\text{txt}}$ steps with a standard ODE solver such as Euler or Heun, producing an estimate $\hat{x}^{\text{text}}_0$ that approximates $x^{\text{text}}_0$. Intuitively, this stage injects semantic reasoning into the canvas by progressively laying down optical traces of language while preserving the global structure of the original image.

\begin{figure}[t]
    \centering
    \scriptsize
    \includegraphics[width=0.9\linewidth]{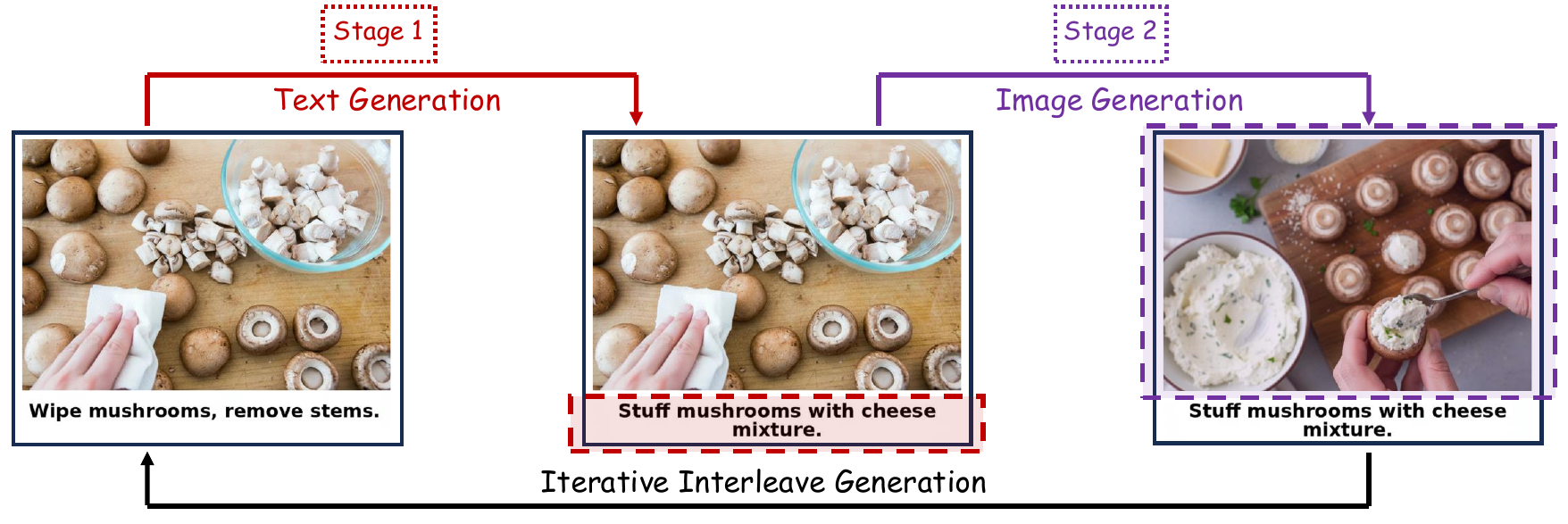}
    \caption{\textbf{Two-stage Canvas Update in \model.} In the first stage, the model ``writes'' the semantic texts on the canvas, while in the second stage, the modal updates the scene image conditioned on the enriched canvas. Iterative interleave generation alternates between these two stages to produce coherent step-wise multimodal sequences.
    }
    \label{fig:two-stage}
\end{figure}

\subsection{Two-Stage Generation}
\label{subsec:two-stage}

Although text and image share the same pixel-native canvas, reasoning (\emph{what} to express) and rendering (\emph{how} to express it) remain conceptually distinct. \model bridges this gap via a two-stage generation process that operates on the same canvas but uses different conditioning signals, as illustrated in Fig.~\ref{fig:two-stage}. Both stages are implemented by a single diffusion backbone $v_{\theta}$ trained end-to-end, ensuring that gradients flow smoothly across reasoning and synthesis.

\paragraph{Stage 1: Visual Reasoning.}
In the first stage, the model writes semantic intent onto the canvas. Starting from the current image $x$ (which can be the initial input or the output of a previous step in a multi-step task), we apply the text-in-image flow matching described in Eq.~\eqref{eq:flow_matching_text}. Integrating the learned velocity field for $T_{\text{txt}}$ steps with the reasoning prompt $c_{\text{reason}}$ yields a reasoning-enhanced latent
\begin{equation}
\hat{x}^{\text{text}}_0 \approx x^{\text{text}}_0,
\end{equation}
where the canvas now contains explicit optical text annotations and structured visual patterns that encode the inferred high-level semantics.

\paragraph{Stage 2: Visual Synthesis.}
In the second stage, the model transforms this reasoning-enhanced canvas into the next scene state. Let $x^{\text{state}}_0$ denote the ground-truth next-state image (for instance, the frame after executing a described action). We again apply flow matching, now with source $x_0 = \hat{x}^{\text{text}}_0$, target $x_1 = x^{\text{state}}_0$, and an execution-oriented context $c_{\text{exec}}$:
\begin{equation}
\mathcal{L}^{\text{state}}_{\text{FM}}
= \mathbb{E}_{t, \hat{x}^{\text{text}}_0, x^{\text{state}}_0}
\left[
\lVert v_{\theta}(x_t, t, c_{\text{exec}}) - (x^{\text{state}}_0 - \hat{x}^{\text{text}}_0) \rVert^2
\right].
\label{eq:flow_matching_visual}
\end{equation}
At test time, we start from $\hat{x}^{\text{text}}_0$ and integrate the velocity field conditioned on $c_{\text{exec}}$ for $T_{\text{vis}}$ steps to obtain the predicted next state $\hat{x}^{\text{state}}_0$.

Because both stages share $v_{\theta}$, the model learns a single canvas dynamics that can both write text and update scene state under different prompts. For multi-step tasks, we simply feed $\hat{x}^{\text{state}}_0$ back as the new $x$ and repeat the two-stage procedure for the next instruction. This yields a sequence of canvases that explicitly record the evolving task state, allowing \model to reason and execute over multiple steps entirely within the visual domain.

\subsection{Regularization with CLIP-Based Alignment}
\label{subsec:regularization}

The flow-matching losses in Eqs.~\eqref{eq:flow_matching_text} and~\eqref{eq:flow_matching_visual} ensure that the model learns consistent transformations between latent states, but they do not directly enforce that the optical text rendered on the canvas faithfully matches the intended linguistic semantics. To strengthen this alignment, we add a CLIP-based contrastive regularization term.

We use a pretrained CLIP model with frozen visual and text encoders, $E^{\text{vis}}_{\phi}$ and $E^{\text{txt}}_{\phi}$. For each training example, we encode the intermediate reasoning canvas $\hat{x}^{\text{text}}_0$ with $E^{\text{vis}}_{\phi}$ and its corresponding text reasoning prompt $c_{\text{reason}}$ with $E^{\text{txt}}_{\phi}$, and normalize both embeddings. A standard CLIP-style contrastive loss is then applied over the mini-batch, encouraging each visual embedding to be closest to its matching text embedding and far from non-matching prompts. We reuse the learned temperature parameter from the original CLIP model and do not fine-tune CLIP itself.

This regularization nudges the model toward rendering optical text that is both legible and semantically faithful, without introducing additional trainable components or heavy computational overhead. In practice, we find that it significantly improves text readability and text-image consistency, especially for longer or more complex on-canvas descriptions.
The full training objective is a weighted sum of the two flow-matching terms and the CLIP-based alignment:
\begin{equation}
\mathcal{L}_{\text{total}} =
\mathcal{L}^{\text{text}}_{\text{FM}} +
\mathcal{L}^{\text{state}}_{\text{FM}} +
\lambda_{\text{CLIP}} \mathcal{L}_{\text{CLIP}},
\label{eq:final_objective}
\end{equation}
where $\lambda_{\text{CLIP}}$ balances the strength of semantic alignment. Together, these components train a single diffusion backbone to reason, write, and render entirely on a shared multimodal canvas, which we later exploit for multi-step tasks by iteratively updating the canvas over instruction sequences.

\subsection{Pre-train with Perceptual Loss}
\label{subsec:pretrain}

Moreover, we observe that current open-source image editing diffusion checkpoints are not capable of generating in-image text. They fail to render legible words and even struggle with text-only images, likely because the absence of certain data during their training. 
We therefore introduce a pre-training stage that teaches the model to write text on the pixel canvas while preserving general visual editing ability. 
Concretely, we formulate text rendering as an image editing task on VQA data. Given an input image $x$ and a question, we construct a target canvas $x^{\text{text}}_0$ by rendering the answer as optical text in a designated region of the image.
We mix these samples with generic image editing data to avoid degrading visual generation quality. 
Since this stage primarily aims to establish reliable optical text writing behavior, we only train one-step input here and leave multi-step rollouts to the subsequent fine-tuning stages.

During pre-training, we add a perceptual alignment loss on the text region to stabilize typography, complementary to the CLIP-based alignment used in fine-tuning. CLIP encourages semantic consistency between prompts and generations, whereas the perceptual loss directly regularizes the visual realization of characters while remaining tolerant to minor spatial shifts. Specifically, we use LPIPS with a frozen VGG backbone $E^{\text{lpips}}_{\phi}$ and compute
$
\mathcal{L}_{\text{perc}}
=
d_{\text{lpips}}\!\left(\mathrm{crop}(\hat{x}^{\text{text}}_0),\, \mathrm{crop}(x^{\text{text}}_0)\right),
$
where $\mathrm{crop}(\cdot)$ extracts the designated in-image text region and $d_{\text{lpips}}(\cdot,\cdot)$ denotes the standard LPIPS metric~\cite{zhang2018perceptual}. This pre-training stage equips the diffusion backbone with foundational text rendering capability on the canvas and provides a strong initialization for downstream multi-step fine-tuning.

%% file: sections/4_experiment.tex
\section{Experiments}
\label{sec:exp}

\subsection{Experimental Setup}
\label{subsec:exp_setup}

\paragraph{\textbf{Benchmarks.}}

We evaluate our method on the \texttt{VSP}~\cite{wu2024vsp} benchmark, which measures spatial planning in a simulated maze-navigation environment. As VSP does not provide a training split, we follow its official data generation recipe to synthesize 4,000 training samples using disjoint maze configurations from official test set,  where evaluation is conducted.
We further assess generalization on two additional datasets with different state-transition structures. The first consists of 2,000 cooking recipes, with interleaved image–text pairs, designed to test multimodal step-wise generation. The second contains 2,000 robot arm trajectories sourced from RLBench~\cite{james2020rlbench}, where we extract key frames and annotate the corresponding action descriptions using GPT-4o with fixed templates. Additional benchmark details, are provided Appx.~\ref{app:benchmark-details}.

\paragraph{\textbf{Baselines.}}

We compare our approach against recent unified multimodal models. Specificallt, we include Anole~\cite{chern2024anole}, finetuning from Chameleon~\cite{team2024chameleon}, MVoT~\cite{li2025imagine}, which supporst both image and text generation, MMaDA~\cite{yang2025mmada}, a diffusion-based unified model capable of text reasoning and image generation, and BAGEL~\cite{deng2025emerging}, a VLM-based unified model with strong multimodal performance. 
For a fair comparison, all baselines are finetuned with the same interleaved action-state sequences and ensure convergence of training. 
For BAGEL, we additionally finetune a text-only variant, where ground truth action sequences are used as supervision. We further report the zero-shot performance of GPT-4o and the finetuned results of \texttt{Qwen2.5-VL-3B}~\cite{bai2025qwen2} for reference. Detailed training configurations and hyper-parameters are specified in Appx.~\ref{app:baseline-details}.

\paragraph{\textbf{Implementation Details.}}

We build \model on top of \texttt{Qwen-Image-Edit 2509} and fine-tune it with LoRA rank 32 using the DiffSynth framework. For pre-training, we mix \texttt{LLaVA-Instruct-150K}~\cite{liu2023llava} for text rendering and a subset of \texttt{OmniEdit-Filtered-1.2M}~\cite{wei2024omniedit} for general image editing, and train on 8 H800 GPUs for approximately 10 days. For step-wise prediction, the model is conditioned only on the most recent image rather than the full history, which reduces training cost with minimal impact on performance. Unless otherwise specified, all experiments use a total batch size of 32 and a fixed random seed of 123. Additional implementation details are provided in Appx.~\ref{app:impl-details}.

\begin{table*}[!t]
    \centering
    \scriptsize
    \caption{\textbf{Comparison Results on VSP tasks.} \model outperforms all relevant baselines. Best result is shown in \textbf{bold} and the second-best result is shown in \underline{underline}.}
    \scriptsize
    \resizebox{0.8\linewidth}{!}{
    \begin{tabular}{@{}lccrrrrr@{}}
        \toprule
        \multirow{2}{*}{Method} & \multirow{2}{*}{Input} & \multirow{2}{*}{Output} & \multicolumn{5}{c}{VSP Planning} \\
        \cmidrule(lr){4-8}
        & & & Grid 3 & Grid 4 & Grid 5 & Grid 6 & Avg. \\
        \midrule
        \multicolumn{8}{c}{(visual-language models)} \\
        Qwen2.5-VL-3B~\cite{bai2025qwen25vl} & \faImage \ ,\ T & T & \underline{0.88} & \textbf{0.82} & \underline{0.73} & \underline{0.47} & \underline{0.72} \\
        GPT-4o Zero-Shot & \faImage \ ,\ T & T & 0.68 & 0.58 & 0.35 & 0.24 & 0.46 \\
        \midrule
        \multicolumn{8}{c}{(unified models)} \\
        Qwen-Image-Edit-2509 & \faImage \ ,\ T & \faImage & - & - & - & - & - \\
        Anole~\cite{chern2024anole} & \faImage \ ,\ T & \faImage \ ,\ T & 0.13 & 0.08 & 0.03 & 0.01 & 0.06 \\
        MVoT~\cite{li2025imagine} & \faImage \ ,\ T & \faImage \ ,\ T & 0.24 & 0.13 & 0.09 & 0.03 & 0.12 \\
        MMaDA~\cite{yang2025mmada} & \faImage \ ,\ T & \faImage \ ,\ T & 0.18 & 0.09 & 0.02 & 0.02 & 0.08 \\
        BAGEL-Text & \faImage \ ,\ T & T & 0.72 & 0.60 & 0.52 & 0.34 & 0.55 \\
        BAGEL-Interleave & \faImage \ ,\ T & \faImage \ ,\ T & 0.32 & 0.23 & 0.26 & 0.13 & 0.24 \\
        \midrule
        \model & \faImage \ ,\ T & \faImage\ (T) & \textbf{0.92} & \underline{0.81} & \textbf{0.79} & \textbf{0.57} & \textbf{0.77} \\
        \bottomrule
    \end{tabular}}
    \label{exp:frozenlake-nav-main}
\end{table*}

\begin{figure}[t]
    \centering
    \scriptsize
    \includegraphics[width=0.7\linewidth]{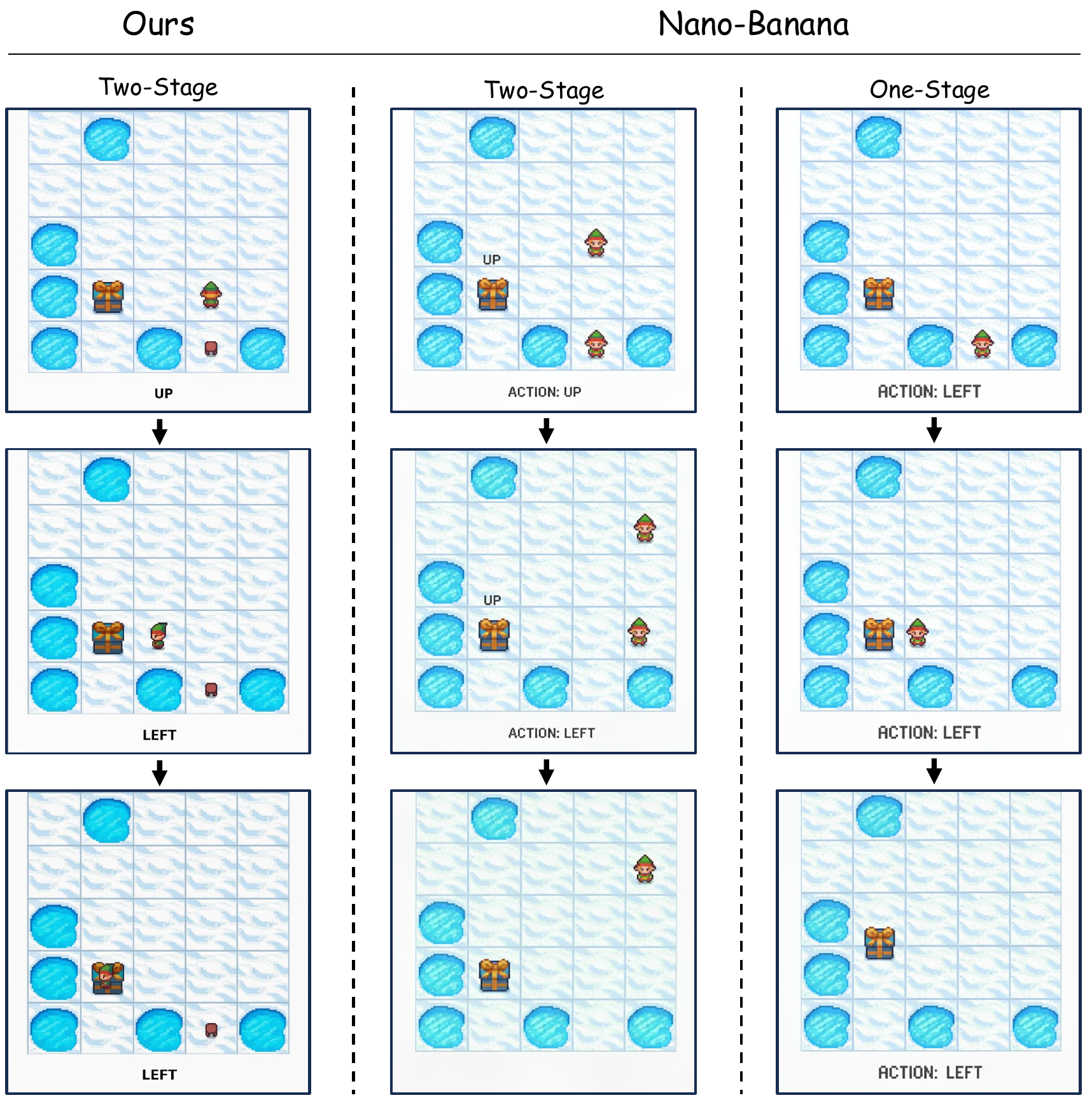}
    \caption{\textbf{Comparison with Nano-Banana on VSP.} Qualitative comparison of action-sequence generation. Our two-stage approach produces more consistent navigation trajectories, while Nano-Banana shows unstable plans and generation in both its one-stage and two-stage variants.}
    \label{fig:comparison-with-nano-banana-frozenlake}
\end{figure}

\subsection{Experimental Results}
\label{subsec:eval}

\paragraph{\textbf{Long-Horizon cross-Modal planning.}}
The VSP planning task serves as a challenging cross-modal reasoning benchmark that requires a model to alternate between interpreting visual states and producing textual actions. While recent unified models aim to bridge language and vision within a single architecture, their ability to maintain spatial consistency and handle long cross-modal transitions remains limited. As shown in Tab.~\ref{exp:frozenlake-nav-main}, most unified models, including Anole~\cite{chern2024anole}, MMaDA~\cite{yang2025mmada}, and even the stronger BAGEL~\cite{deng2025emerging}, struggle to generate accurate interleaved action–state sequences, with average success rates below 0.25. 
We attribute this weakness to the limited visual generation fidelity of unified models. The resulting visual errors quickly accumulate across steps, causing the predicted trajectories to collapse after only two or three updates.
The improved performance of text-only BAGEL further confirms that these models reason better when visual synthesis is removed from the pipeline, indicating that they still rely heavily on the underlying text model's reasoning capability.
In contrast, our \model maintains coherent visual trajectories and stable multi-step reasoning across all grid configurations, achieving a 0.77 average overall success rate, substantially outperforming all baselines. These results indicate that explicitly modeling transformation-conditioned visual updates within a shared canvas can significantly improve long-horizon cross-modal reasoning.

Additionally, we also compare against the autoregressive Qwen2.5-VL model. Even when finetuned on the same data, its performance remains lower than our model, by 0.05 on average. The limited zero-shot performance of GPT-4o further highlights the difficulty of long-horizon multimodal state tracking. Together, these results suggest the potential of modeling explicit visual state transitions through our text-in-image generation paradigm.
More results and the comparison of inference efficiency are provided in Appx.~\ref{app:inference-speed}.

\begin{wrapfigure}{r}{0.6\textwidth}
    \centering
    \scriptsize
    \includegraphics[width=\linewidth]{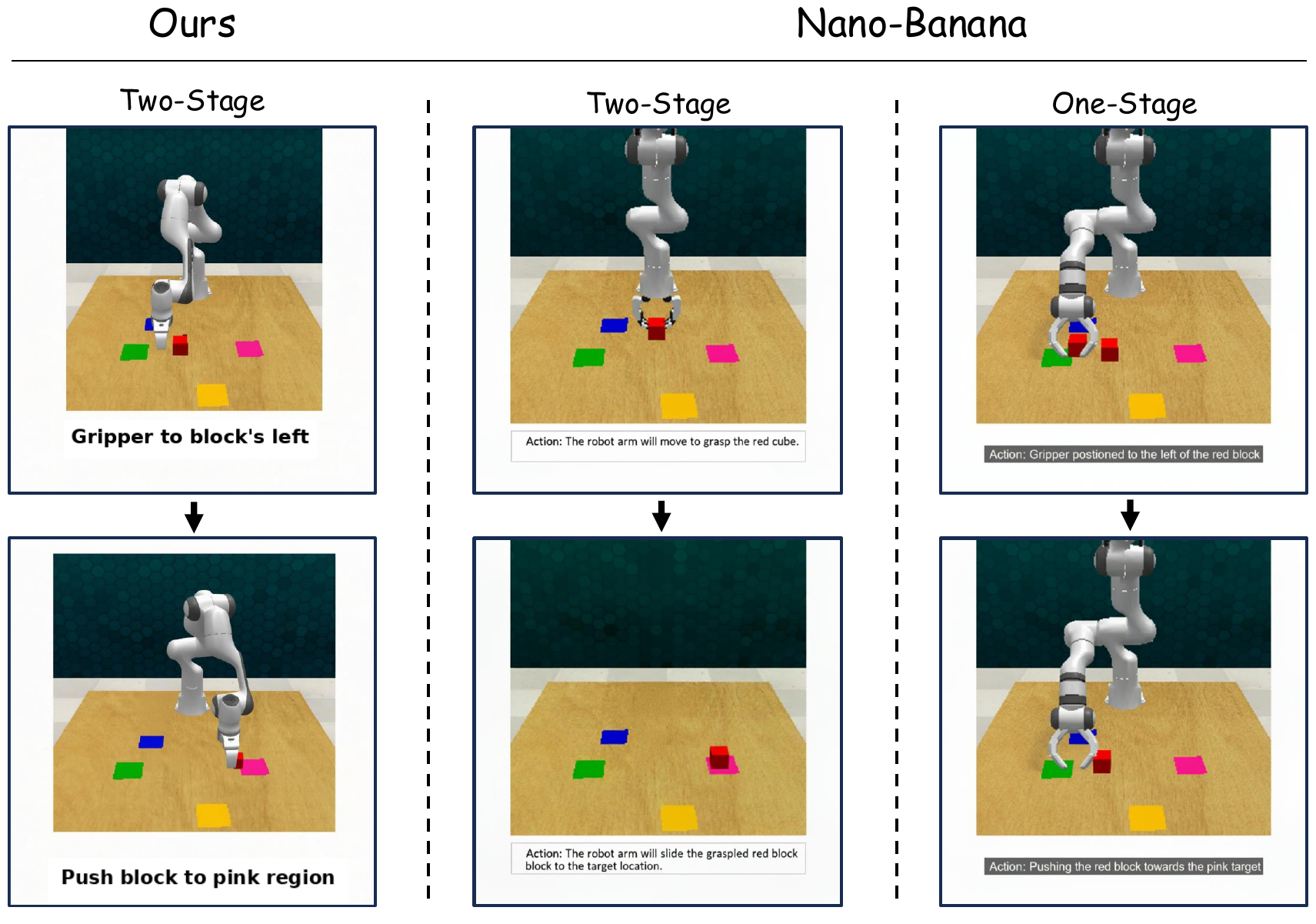}
    \caption{\textbf{Comparison with Nano-Banana on RLBench.} Qualitative comparison of action-sequence generation. Our model yields clearer action intent and more accurate scene transitions, while both Nano-Banana variants often produce ambiguous or inconsistent results.}
    \label{fig:comparison-with-nano-banana-rlbench}
\end{wrapfigure}

\paragraph{\textbf{Qualitative comparisons on visual quality.}}
To further assess the qualitative behavior of our model, we compare \model with  
\texttt{Nano\hyp{}Banana}.
In Fig.~\ref{fig:comparison-with-nano-banana-frozenlake}, we evaluate Nano-Banana under both the one-stage and two-stage settings described in Sec.~\ref{sec:method}, with detailed prompts provided in Appx.~\ref{app:impl-details}. Although Nano-Banana is evaluated in a zero-shot manner, the comparison still provides substantial evidence of the effectiveness of our model. \model consistently produces coherent text-in-image instructions and accurate next-state predictions throughout the entire trajectory, ultimately guiding the agent to the target cell. In contrast, Nano-Banana exhibits unstable and often incorrect visual transitions, frequently hallucinating duplicated agents, missing agents, or inconsistent treasure locations. Even when the two-stage setting encourages it to generate more legible text, the resulting next-state frames often fail to match the stated actions, revealing a collapse in cross-modality consistency. These observations further highlight that \model produces stable, logically aligned canvas sequences where embedded text and visual state transitions evolve coherently, demonstrating genuine cross-modal reasoning rather than independent text- or image-only heuristics.

Beyond grid-based navigation, we also provide qualitative results on the synthesized RLBench dataset. As shown in Fig.~\ref{fig:comparison-with-nano-banana-rlbench}, similar to the observations above, Nano-Banana again suffers from severe hallucinations, including disappearing robot arms, duplicated manipulated objects, and its next-state predictions remain inconsistent with the optical text it generates under both one-stage and two-stage settings. Although without an explicit numerical metric, the pervasive visual instability and persistent misalignment between predicted actions and resulting states transitions underscore Nano-Banana's limitations in coherent cross-modal reasoning. In contrast, \model consistently produces physically plausible visual transitions that follow the embedded action descriptions, demonstrating strong performance across domains and reinforcing the robustness of our text-in-image paradigm.

\begin{table}[t]
    \centering
    \caption{\textbf{Eval. of visual quality.} Quantitative evaluation of DINO and CLIP scores on Recipe and RLBench. \model achieves consistently strong visual fidelity under image–text, image-only, and text-only conditioning settings.}
    \scriptsize
    \setlength{\tabcolsep}{4pt}
    \begin{tabular}{@{}lcccccc@{}}
        \toprule
        & \multicolumn{3}{c}{DINO Score} & \multicolumn{3}{c}{CLIP Score} \\
        \cmidrule(lr){2-4} \cmidrule(lr){5-7}
        & \faImage \ ,\ T & \faImage & T & \faImage \ ,\ T & \faImage & T \\
        \midrule
        Recipe & 56.01 & 58.53 & 53.24 & 79.95 & 80.09 & 77.68 \\
        RLBench & 80.72 & 81.67 & 73.06 & 89.79 & 89.99 & 91.93 \\
        \bottomrule
    \end{tabular}
    \label{exp:visual-quality}
\end{table}
\paragraph{\textbf{Quantitive analysis on visual quality.}}
We further assess the visual fidelity of our generated outputs using perceptual similarity metrics. As shown in Table~\ref{exp:visual-quality}, we report DINO and CLIP similarity scores, where higher values indicate better perceptual alignment with ground truth reference images. To provide a fine-grained analysis, we separately evaluate the entire text-embedded image (\faImage, T), the pure visual region (\faImage), and the pure text region (T). Across both the Recipe and RLBench datasets, our model achieves consistently strong visual similarity across all evaluation regions, demonstrating robustness in generating coherent visual content alongside accurate textual overlays. The lower similarity scores observed in the Recipe domain can be attributed to the intrinsic diversity of this task, where a single input may map to multiple valid outputs, naturally reducing similarity-based metrics. In contrast, the robotics dataset features a single deterministic next state, allowing models to edit directly from the current visual frame, thereby producing higher similarity scores. Despite this discrepancy, the model maintains high visual quality across both domains, further confirming the effectiveness of our text-in-image design for unified visual–textual generation.

These results align with our \textbf{text-in-image} generation paradigm. By projecting textual instructions directly into the visual domain, our model performs all intermediate reasoning within a unified visual representation. In contrast, previous unified models repeatedly switch between text and image embeddings during multi-step navigation. Such cross-modality transitions amplify representation mismatches across long horizons, causing accumulative errors and eventual prediction collapse, especially in larger grids. In contrast, our text-in-image generation paradigm eliminates this repeated modality switching, alleviating the error accumulation and thus enabling stable long-horizon multimodal generation.

Overall, these findings demonstrate that performing all reasoning within a unified visual modality is crucial for robust, long-horizon decision making. Prior unified models struggle in large visual state spaces, and VLMs deteriorate rapidly due to multi-step cross-modality inconsistencies that accumulate over time. Nano-Banana further highlights this limitation through pervasive hallucinations and action–state mismatches across both FrozenLake and RLBench, even when the generated optical text appears legible. In contrast, our method avoids these pitfalls entirely through the proposed \textbf{text-in-image} mechanism, which eliminates repeated transitions between text and image domains and prevents error accumulation over long action sequences. The minimal performance drop observed in larger grids, combined with strong qualitative stability and high perceptual similarity across diverse datasets, validates the robustness and generalization capability of our approach in increasingly challenging navigation and visuomotor settings.

\subsection{Analysis}
\label{subsec:qual}

\begin{table}[t]
    \centering
    \caption{\textbf{Ablation study of key components.} Removing either the CLIP-based loss or the two-stage training mechanism degrades planning accuracy across grid sizes.}
    \scriptsize
    \setlength{\tabcolsep}{4pt}
    \resizebox{0.8\linewidth}{!}{
    \begin{tabular}{@{}lrrrrr@{}}
        \toprule
        & Grid 3 & Grid 4 & Grid 5 & Grid 6 & Avg. \\
        \midrule
        \textbf{\model} & 0.92 & 0.81 & 0.79 & 0.57 & 0.77 \\
        \midrule
        (-) Text-in-Image Pretraining & 0.48 & 0.26 & 0.12 & 0.04 & 0.23 \\
        (-) CLIP-based Loss & 0.78 & 0.68 & 0.42 & 0.30 & 0.55 \\
        (-) Two-Stage & 0.66 & 0.56 & 0.30 & 0.18 & 0.43 \\
        \bottomrule
    \end{tabular}
    }
    \label{exp:frozenlake-nav-ablation}
\end{table}

\paragraph{\textbf{Ablation study of key components.}}
To better understand the contribution of each component in our model, we report ablation results in Tab.~\ref{exp:frozenlake-nav-ablation}. The full pipeline achieves the best performance across all grid settings, which validates the importance of both components.
Firstly, without the large-scale text-in-image pretraining, the model struggles to consistently generate valid on-canvas text, and early text errors quickly propagate to incorrect state updates. Initializing from pretrained checkpoints lays a strong foundation for reliable text rendering and yields clear gains.
The CLIP-based alignment loss provides explicit semantic supervision for the rendered in-image text, which is otherwise only weakly constrained by the flow-matching objectives. Since the optical text occupies a small region of the canvas, the model can partially ignore it or produce misaligned and unreadable characters without noticeably affecting the global denoising loss. Removing this therefore degrades text legibility and consistency, which in turn lowers navigation success across grid sizes

Without the two-stage framework, the model must render the action text and synthesize the next-state image in one step, coupling decision making with visual state updates. Separating action rendering from state generation simplifies learning and improves robustness. Consequently, removing the two-stage design consistently reduces success rates, especially on larger grids. Overall, the ablations indicate that large-scale pretraining, CLIP-based alignment, and the two-stage decomposition are complementary, and their combination is crucial for reliable in-image text generation and multi-step execution.

\begin{table}[t]
    \centering
    \caption{\textbf{Evaluation of action prediction with ground truth simulation.} Both equipped with oracle state visualization, \model performs competitively with strong VLMs, showcasing the potential of text-in-image generation paradigm.}
    \scriptsize
    \setlength{\tabcolsep}{4pt}
    \begin{tabular}{@{}lrrrrr@{}}
        \toprule
        & Grid 3 & Grid 4 & Grid 5 & Grid 6 & Avg. \\
        \midrule
        \textbf{\model} & 0.97 & 0.87 & 0.83 & 0.69 & 0.84 \\
        \midrule
        Qwen2.5VL-3B & 0.94 & 0.85 & 0.80 & 0.65 & 0.82 \\
        Qwen2.5VL-7B & 0.99 & 0.92 & 0.88 & 0.76 & 0.89 \\
        \bottomrule
    \end{tabular}
    \label{exp:frozenlake-nav-action-comparison}
\end{table}

\paragraph{\textbf{Effect of ground truth visual inputs.}}
To further understand how visual editing error accumulation affects downstream performance, we conduct an ablation where we replace the model-generated next-state canvas with the ground-truth environment simulation output, while still using the action predicted by our text-generation stage. 
This removes compounding synthesis noise across steps and evaluates decision quality under the same step-wise setting as the Qwen2.5-VL baselines.
As shown in Tab.~\ref{exp:frozenlake-nav-action-comparison}, ground-truth visual input consistently boosts success across all grid sizes, both in our model and VLMs, exposing a stronger upper bound of \model when decoupled from rendering errors. It surpasses \texttt{Qwen2.5-VL 3B} and becomes competitive with \texttt{Qwen2.5-VL 7B}.
Overall, these results suggest that, within a fixed action space, performance can benefit from cleaner visual state updates, pointing to the potential of our text-in-image paradigm.

\begin{wraptable}{r}{0.4\textwidth}
    \centering
    \caption{\textbf{Comparison on general reasoning tasks.} \model performs competitively with strong VLMs, showcasing the potential of our paradigm.}
    \scriptsize
    \setlength{\tabcolsep}{4pt}
    \resizebox{\linewidth}{!}{
    \begin{tabular}{@{}lrr@{}}
        \toprule
        & CocoQA & Visual7W \\
        \midrule
        Qwen2.5VL-3B & 0.72 & 0.86 \\
        \textbf{\model} & 0.58 & 0.78 \\
        \bottomrule
    \end{tabular}
    }
    \label{exp:general-reasoning-comparison}
\end{wraptable}

\paragraph{\textbf{Evaluation on general reasoning tasks.}}
Beyond producing actions in a fixed action space, we further probe the model's general textual reasoning capability. In this setting, instead of interleaving in-image text with image editing on the canvas, \model directly produces textual answers conditioned on the query. We randomly sample 200 test examples for evaluation. For Visual7W, we additionally continue finetuning on the training split to adapt the model to the multiple-choice answering format. As shown in Tab.~\ref{exp:general-reasoning-comparison}, \model lags behind general reasoning VLMs but remains competitive given the gap in supervision. Models such as \texttt{Qwen2.5-VL} benefit from large-scale instruction tuning, whereas \model is only exposed to limited textual supervision from LLaVA data. More qualitative examples are in the appendix.

\paragraph{\textbf{Limitation and future work.}}

While the results above provide encouraging evidence for the proposed text-in-image paradigm, there remains room for improvement.
First, \model is less reliable on long-form reasoning that requires producing multi-sentence explanations. This is expected given that the underlying editing backbone is primarily optimized for image synthesis and localized text rendering, rather than rich free-form text generation. 
In practice, we find that decomposing long outputs into sequential steps improves stability, suggesting a promising direction for scaling long-horizon reasoning by structuring generation over multiple canvas updates.
Second, although \model can generate and edit images as part of the reasoning process, visual synthesis errors can still occur, and occasional inconsistencies in the rendered state may accumulate over long trajectories. We include representative failure cases in the appendix.

More broadly, our goal is to establish the feasibility of a diffusion-based unified framework where reasoning and generation are carried out on a shared, writable canvas. Despite current limitations, the empirical results already demonstrate stable interleaved generation across diverse domains. We anticipate that these limitations can be substantially mitigated with larger-scale training and stronger supervision. Overall, we view \model as an initial step toward a new and promising paradigm for unified multimodal generation and reasoning.

%% file: sections/5_conclusion.tex
\section{Conclusion}

In this work, we introduce \model, a diffusion-based unified model that renders both text and images directly within a shared pixel canvas. By treating linguistic and visual content as a single visual modality, \model achieves multimodal generation without relying on discrete text tokens or separate decoding heads. With coherent text-in-image joint generation, \model naturally integrates multi-step reasoning with image synthesis, and produces interpretable canvas trajectories that explicitly track task progress. Experimental results demonstrate that \model consistently outperforms prior unified models. These highlight the potential of canvas-based multimodal generation as a promising paradigm for unified models within a single architecture.

%% file: sections/a_supplement.tex
\setcounter{section}{0}
\title{UniCanvas: A Diffusion-based Unified Model \\ for Text-in-Image Joint Generation}

\begin{center}
\textbf{{\Large
UniCanvas: A Diffusion-based Unified Model
\\[5pt]
for Text-in-Image Joint Generation}}
\\[7pt]
{Supplementary Material}
\end{center}

\section{Experimental Details}
\label{app:exp-details}
As outlined in Section 4.1 of the main paper, we evaluate \model on three distinct datasets requiring multi-step reasoning.

\subsection{Benchmark Details}
\label{app:benchmark-details}

\paragraph{VSP.}
The VSP benchmark evaluates spatial planning in a maze navigation environment. Following the official data generation recipe, we synthesize 1{,}000 training trajectories with grid sizes between $3\times3$ and $6\times6$, random start and goal locations, and randomly sampled obstacle layouts with less than 30\% blocks. Each trajectory is converted into a sequence of state and action pairs that serve as supervised training samples. Models are trained solely on the synthesized split, and all results are reported on the official VSP main-task test set. 

To support our interleaved design, we further augment the original training data in VSP by generating text-in-image and next-state image variants. \textbf{Text-in-image}: For each state image, we embed the corresponding action label (\eg, ``UP'', ``RIGHT'') in the bottom region of the image. \textbf{Next-state image}: The paired next-state frame is similarly processed by adding the same action text, ensuring both images in a state–action pair contain consistent action cues. This augmentation produces an aligned interleaved sequence of visual states and action-annotated images, allowing the model to jointly learn spatial reasoning and action-conditioned transitions. Finally, we provide the prompts used during training and inference. The exact prompt template is shown in Table~\ref{tab:prompt-vsp}.

\paragraph{Recipe.} For the Recipe benchmark, we manually curate a dataset of 2,000 cooking episodes, each consisting of step-by-step image–text pairs. Each episode corresponds to a complete dish and is associated with a sequence of short instructional steps. At each step, the model is given the current visual frame together with the textual history of previously completed instructions. The task is to generate the next instruction, which is appended as text at the bottom of the image, and then to produce the updated image reflecting the newly executed step. We randomly split the 2,000 episodes into 1,800 training instances and 200 test instances. The exact prompt template used during both training and inference is provided in Table~\ref{tab:prompt-recipe}.

\paragraph{RLBench.}
We construct with 1{,}000 robot arm trajectories sourced from RLBench. For each trajectory, we extract a small number of key frames and use GPT-4o to annotate the corresponding high-level action descriptions. The dataset is randomly split into training and test sets with a 9:1 ratio. The prompt we used is demonstrated in the Table~\ref{tab:prompt-rlbench}.

\paragraph{General Visual Reasoning.}
To further assess the model’s capability on general visual reasoning tasks, we evaluate on COCO-QA and Visual7W, which represent open-ended and multiple-choice visual question answering settings, respectively. We randomly select 200 samples from each dataset for evaluation. The detailed prompt used during inference can be found in Table~\ref{tab:prompt-vqa}.
In addition, we also provide our prompt template for the long context attempt discussed in the main paper. Specifically, for long answers we split the target response into sentence-level chunks and generate them sequentially. At each step, we apply OCR to the previously generated in-image text and feed the extracted text back as the ``\texttt{Current Answer}'' context, asking the model to continue writing the next chunk on the canvas. This design reduces the difficulty of producing long, coherent text in a single pass and allows the model to iteratively extend its answer.

\begin{table}[t]
    \centering
    \footnotesize
    \caption{\textbf{Prompt of the VSP task.} The task instructions include the required textual action description and the corresponding image update behavior. The table details both the text generation prompt and  the image generation prompt.}
    \scriptsize
    \setlength{\tabcolsep}{4pt}
    \renewcommand{\arraystretch}{1.05}
    \begin{tabular}{p{0.95\linewidth}}
        \toprule
        \textbf{Text Generation Prompt:} \\[2pt]
        Determine the action to take to move the elf one block toward the treasure in the grid map. The elf moves exactly one block in one of four possible directions: UP, DOWN, LEFT, or RIGHT. Add the action text at the bottom of the image. Keep the grid state completely unchanged. Only add the text label at the bottom indicating the action to take. \\[4pt]
        \midrule
        \textbf{Image Generation Prompt:} \\
        Execute the action shown in the text at the bottom of the image. Move the elf one block in the specified direction (UP, DOWN, LEFT, or RIGHT). Output the updated grid state with the elf in its new position. \\
        \bottomrule
    \end{tabular}
    \label{tab:prompt-vsp}
\end{table}

\begin{table}[t]
    \centering
    \caption{\textbf{Prompt of the Recipe task.} The specific task instructions, including the recipe name and the current steps, are labeled with \texttt{typewriter}.}
    \scriptsize
    \setlength{\tabcolsep}{4pt}
    \renewcommand{\arraystretch}{1.05}
    \begin{tabular}{p{0.95\linewidth}}
        \toprule
        \textbf{Text Generation Prompt:} \\[2pt]
        Write a step-by-step cooking recipe for \texttt{Homemade Applesauce}. The current image shows the current stage. Some steps: \texttt{(1) Add apples, lemon, cinnamon, sugar, water, salt..}, have already been completed. Reason about the next cooking step needed to continue the recipe. Then, write this next instruction as text at the bottom of the image. Keep the visual content of the image completely unchanged and only add the new step instruction at the bottom. \\[4pt]
        \midrule
        \textbf{Image Generation Prompt:} \\
        Write a step-by-step cooking recipe for \texttt{Homemade Applesauce}. The image shows the current state of the dish with the next cooking step written as text at the bottom. Update the image to reflect the scene after executing this step. The visual change should follow the bottom text instruction, and the text itself should remain at the bottom. \\
        \bottomrule
    \end{tabular}
    \label{tab:prompt-recipe}
\end{table}

\begin{table}[t]
    \centering
    \caption{\textbf{Prompt of the RLBench task.} The specific task instruction is labeled with \texttt{typewriter}.}
    \scriptsize
    \setlength{\tabcolsep}{4pt}
    \renewcommand{\arraystretch}{1.05}
    \begin{tabular}{p{0.95\linewidth}}
        \toprule
        \textbf{Text Generation Prompt:} \\[2pt]
        This is a robotic manipulation task: \texttt{slide the block to the pink target}. Given the current scene, reason about the next action step required to accomplish this goal. Then, add the action description text at the bottom of the image. Keep the scene state completely unchanged. \\[4pt]
        \midrule
        \textbf{Image Generation Prompt:} \\
        This is a robotic manipulation task: \texttt{slide the block to the pink target}. The image shows the current scene with an action description at the bottom. Update the image to show the scene after executing this action. The robot should move according to the action text. \\
        \bottomrule
    \end{tabular}
    \label{tab:prompt-rlbench}
\end{table}

\begin{table}[t]
    \centering
    \caption{\textbf{Prompt templates for General Visual Reasoning Tasks.} We additionally include the template for our long context attempt, where the model generates the answer sequentially in sentence-level chunks. The task-specific instruction is highlighted in \texttt{typewriter}.}
    \scriptsize
    \setlength{\tabcolsep}{4pt}
    \renewcommand{\arraystretch}{1.05}
    \begin{tabular}{p{0.95\linewidth}}
        \toprule
        \textbf{Text Generation Prompt:} \\[2pt]
        Question: \{\texttt{Question}\} \\
        In-Image Text Answer: \\[4pt]
        \midrule
        \textbf{Long Text Generation Prompt:} \\[2pt]
        Question: \{\texttt{Question}\} \\
        Current Answer: \{\texttt{Previous Output}\} \\
        Continue the In-Image Text Answer: \\
        \bottomrule
    \end{tabular}
    \label{tab:prompt-vqa}
\end{table}

\begin{figure*}[t]
    \centering
    \includegraphics[width=1\linewidth]{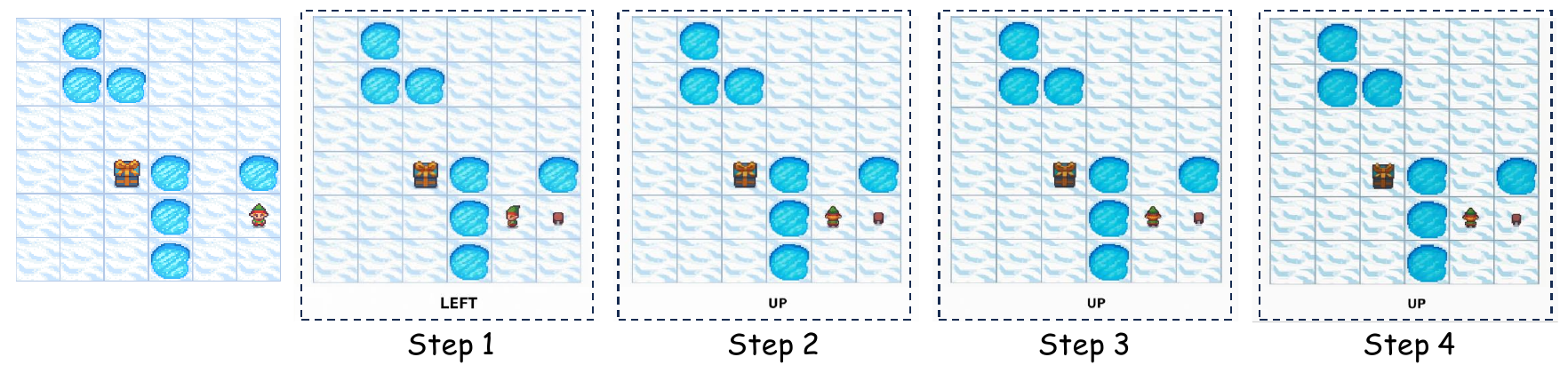}
    \caption{\textbf{Failure case of long-horizon rollouts.} \model becomes unreliable at predicting the correct next-state image when evaluated in the long sequence scenarios. Once the predicted state deviates, subsequent rollouts follow an incorrect trajectory.}
    \label{fig:failure1}
\end{figure*}

\subsection{Baseline Details}
\label{app:baseline-details}

\paragraph{Anole}
We finetune the \texttt{ Anole-7B} model with the officially released code (\url{https://github.com/GAIR-NLP/anole}). Results are reported with the best hyper-parameters (learning rate, epochs, e.t.c.) searched in 10 runs, training with the same data as ours.

\paragraph{MVoT}
We reproduce the results following the original paper~\cite{li2025imagine} and released code (\url{https://github.com/chengzu-li/MVoT}). For fairness, we finetune the base model (\texttt{Anole-7B}) using the same data as in our framework in an action and image interleaved format, following their recipe with hyperparameters stated in \texttt{MVoT} paper and used in the repo. 

\paragraph{MMaDA}
We evaluate the \texttt{MMaDA-8B} model as a unified diffusion baseline, using the official implementation and released checkpoint (\url{https://github.com/Gen-Verse/MMaDA}). We finetune the model on our action–image interleaved data with the same training recipe as in our framework. Since the original MMaDA setup does not natively support image editing, we reuse the image tokenizer provided for their multimodal understanding pipeline.

\paragraph{BAGEL}
For \texttt{BAGEL-7B-MoT}, we build on the official codebase (\url{https://github.com/ByteDance-Seed/Bagel}). We finetune it on both the same action–image interleaved format and the text-only format.

\subsection{Implementation Details}
\label{app:impl-details}

\paragraph{Architecture.} Our system is constructed on top of \textit{Qwen-Image-Edit-2509}, which serves as the underlying diffusion backbone and adopts a Diffusion Transformer (DiT) architecture. To process visual inputs, we employ a frozen variational autoencoder used in \textit{Qwen2.5-VL}, which supports diverse resolutions. Text-based conditioning for both the reasoning prompt ($c_{\text{reason}}$), and the execution prompt ($c_{\text{exec}}$) is handled through the tokenizer used in \textit{Qwen2.5-VL}, which ensures robust alignment between textual instructions and the visual generative process.

\paragraph{Training Hyperparameters.} We train the model using the DiffSynth framework (\url{https://github.com/modelscope/DiffSynth-Studio}) on a cluster equipped with 8 NVIDIA H100 GPUs. Training uses a total batch size of 32 and applies the AdamW optimizer with $\beta_1=0.9$ and $\beta_2=0.999$. The learning rate remains fixed at $1 \times 10^{-4}$ throughout the entire training schedule. For the semantic alignment objective described in Eq.~\ref{eq:final_objective}, we set the CLIP-based regularization weight $\lambda_{\text{CLIP}} = 0.1$ to balance visual fidelity against consistency with the textual conditions.

\paragraph{Inference Settings.} At inference time, we adopt a \textbf{Two-Stage Generation} procedure that decouples the reasoning and synthesis process. In the first stage, corresponding to visual reasoning, the model performs $T_{\text{txt}}=40$ denoising steps using the Euler ODE solver to generate a text-annotated intermediate canvas that reflects the reasoning output. In the second stage, corresponding to visual synthesis, the system performs another $T_{\text{vis}}=40$ Euler ODE steps. This stage updates the visual content according to the text instruction shown at the bottom of the canvas and produces the next visual state that aligns with the specified execution step.

\section{Analysis and Case Study}
\label{app:case-study}

\begin{figure*}
    \centering
    \includegraphics[width=1\linewidth]{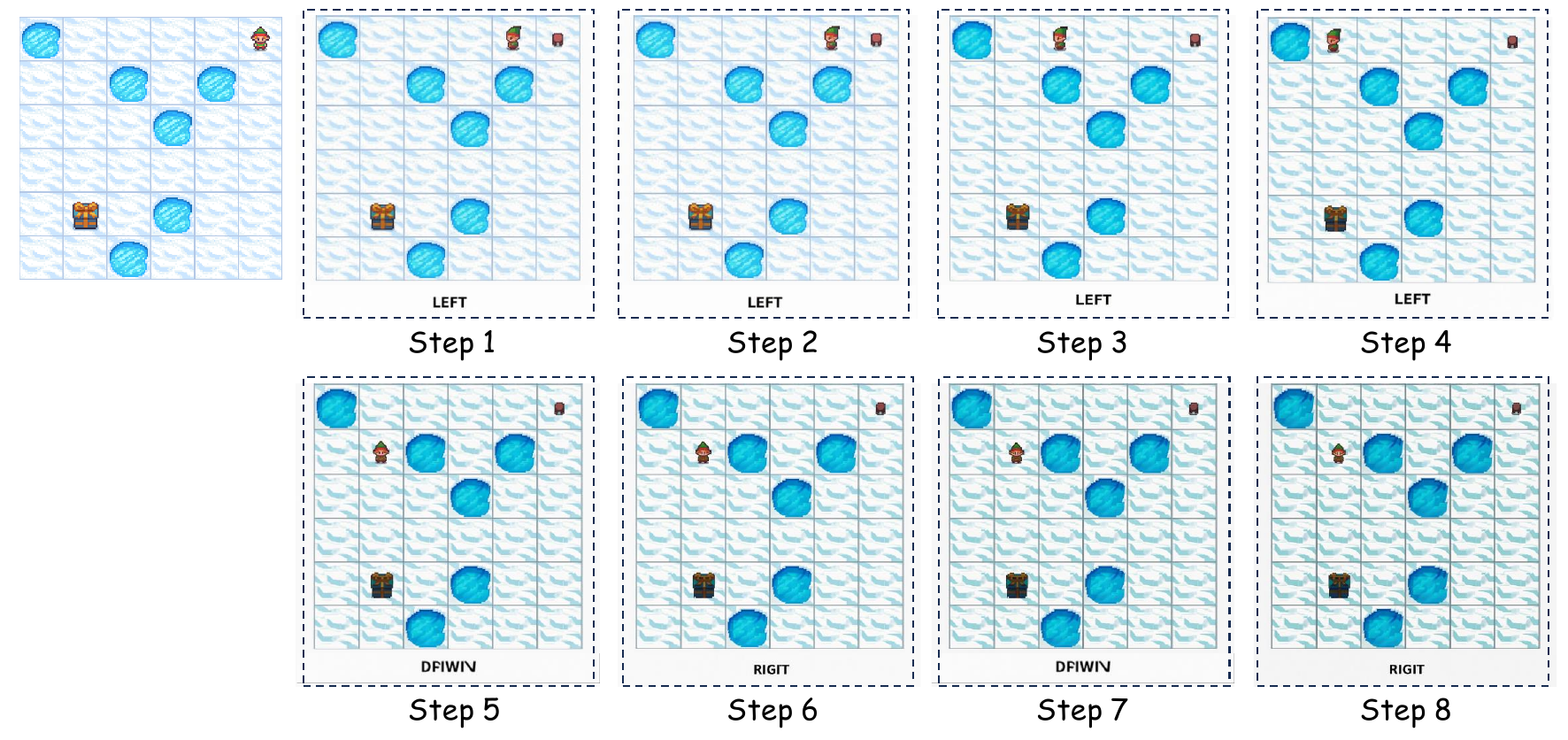}
    \caption{\textbf{Failure case of text-in-image generation.}  \model fails to produce correct text-in-image renderings in long sequence scenarios. As action tokens become misinterpreted, the predicted states drift from the true dynamics, causing subsequent rollouts to follow an inconsistent and increasingly incorrect trajectory.}
    \label{fig:failure2}
\end{figure*}

\begin{figure*}[t]
    \centering
    \includegraphics[width=0.8\linewidth]{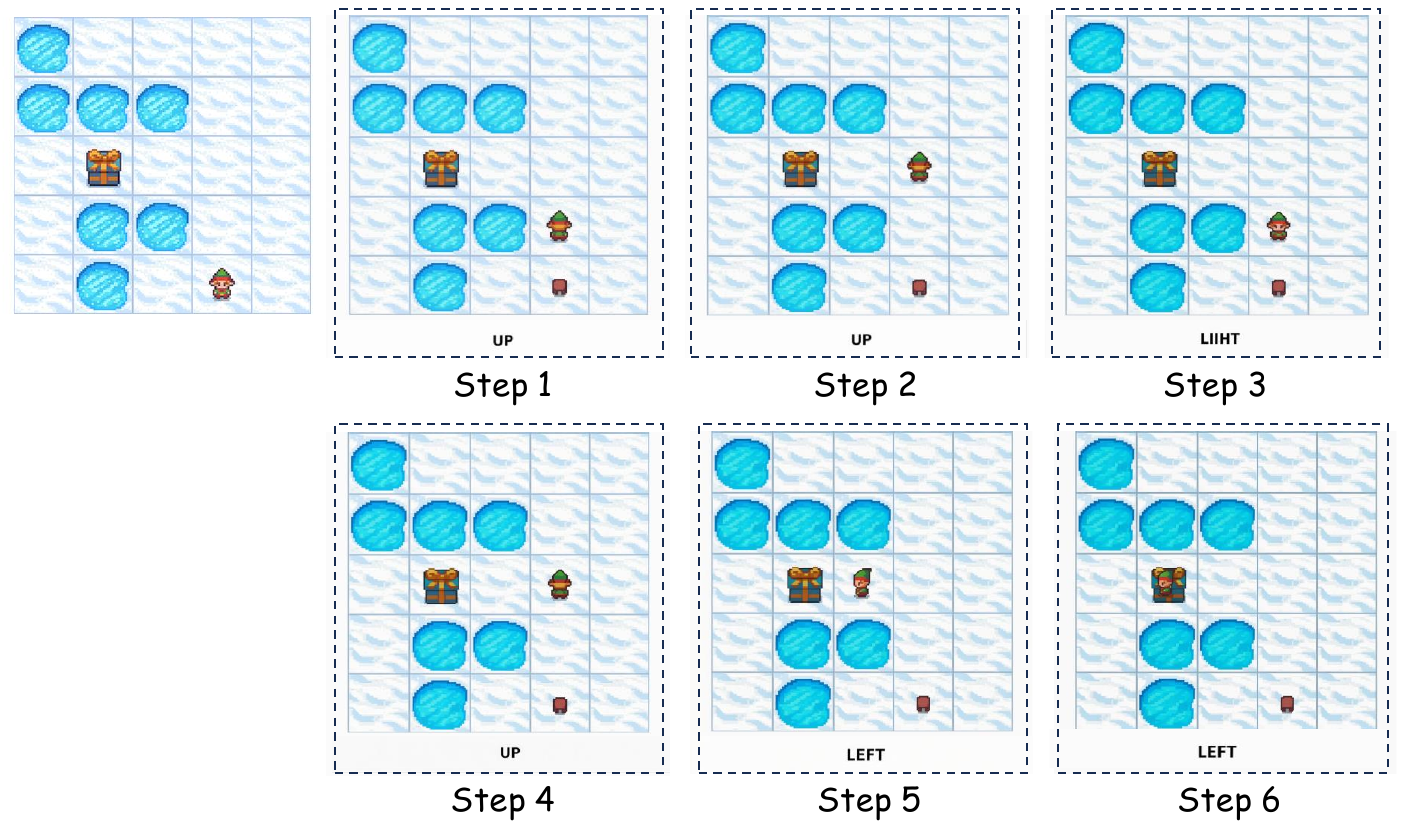}
    \caption{\textbf{Self-correction during long-horizon rollouts.} \model initially predicts an incorrect text-in-image action, causing its next-state prediction to drift from the true dynamics. However, the model is able to recover from this mistake and realign its subsequent predictions, ultimately guiding the agent to the correct goal state.}
    \label{fig:success1}
\end{figure*}

\subsection{Failure Case of State Image Generation}
\label{app:failure-image-gen}
A primary failure mode of our approach emerges as the prediction horizon increases. While \model can reliably generate coherent text-in-image signals and accurate next-state predictions during early or shallow timesteps, the quality gradually degrades as the sequence progresses, where the generation errors inevitably accumulate. 

As illustrated in Fig.~\ref{fig:failure1}, this degradation becomes particularly evident under the most challenging $6 \times 6$ grid setting. In long-horizon rollouts, \model is able to synthesize correct text-in-image renderings, yet it increasingly struggles to predict accurate next-state images. Once the predicted state deviates, the rollout becomes trapped in an incorrect trajectory, ultimately propagating compounding reasoning errors.

\subsection{Failure Case of Action Text Generation}
\label{app:failure-image-gen}

Beyond the above issue, we identify a second failure mode highlighted in Fig.~\ref{fig:failure2}, where the opposite phenomenon occurs, the model predicts incorrect text-in-image signals while its state predictions remain plausible. In this case, an early error in the rendered optical text, such as the incorrect symbol shown in the second row, second column, causes the model to repeatedly propagate this mistake. The erroneous agent state continues to appear in the same position for the remainder of the rollout, which indicates that the model has stuck onto an incorrect textual representation. Moreover, the misrendered optical text often does not correspond to any text in our dataset and may not even contain visually corrupted or unknown characters, further degrading the semantic consistency of the sequence.

In the latter portion of long-horizon rollouts for both failure types, we observe that the rendered optical text becomes noisy or unreadable, and the predicted environment states may drift from the ground truth. In severe cases, the model hallucinates entirely incorrect layouts or nonsensical textual content, causing the generated sequence to lose coherence altogether.

We attribute these behaviors to cumulative error propagation inherent in the sequential inference pipeline. Each inference step, comprising both the generation of the text-in-image signals and the prediction of the next-state image, is executed independently, conditioned only on the previously generated state. As a result, any slight deviation from the true transition dynamics or small visual artifacts can be amplified at subsequent timesteps. When sequence length increases, such accumulated deviations lead to pronounced errors in both optical text and environment representations.

\subsection{Case of Self-Correction}
\label{app:success-self-correction}
In Fig.~\ref{fig:success1}, we present a contrasting case that reveals an emergent form of self-correction. In step 3, the model generates an incorrect text-in-image, ``LIIHT'', which is not even a valid word. Due to this error, the model subsequently predicts an incorrect next-state image where the agent moves downward. Despite this hallucination, the model is capable of recovering. It resumes inference based on the misaligned state and eventually stabilizes its predictions. Remarkably, even after diverging from the correct trajectory, the model ultimately reaches the goal by reestablishing coherent text-in-image and next-state predictions. This example demonstrates that \model, while susceptible to local prediction errors, can self-correct when provided sufficient contextual continuity within the environment.

\subsection{Quantitative Results on General Visual Reasoning}
\label{app:llava-text-gen}

\begin{figure*}[t]
    \centering
    \includegraphics[width=1.0\linewidth]{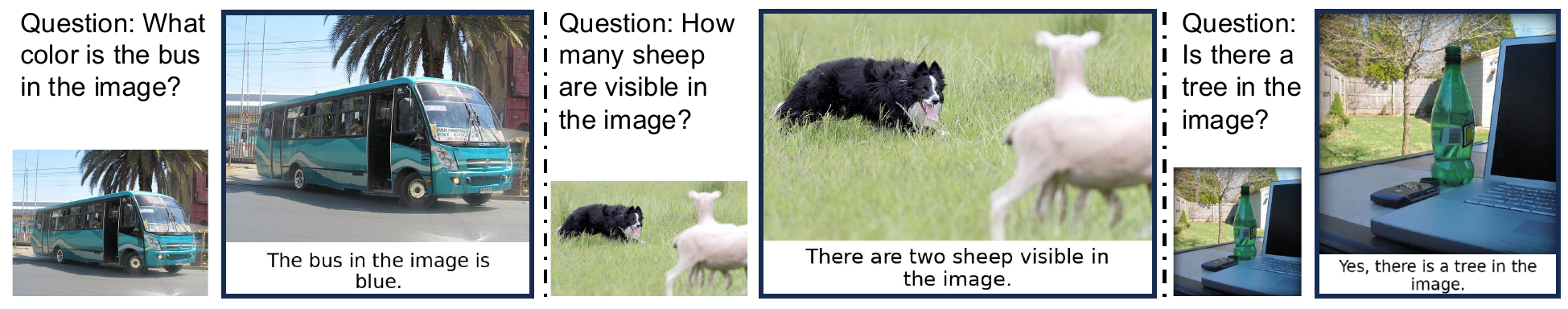}
    \caption{\textbf{Quantitative results on general visual reasoning.} \model directly renders textual answers on the canvas.}
    \label{fig:llava_case}
\end{figure*}

We present quantitative results on general visual reasoning in Fig.~\ref{fig:llava_case}.
With the pretraining, \model produces reliable answers on open-ended questions by directly rendering the textual response on the canvas.
The examples show that the model can correctly identify objects, attributes, and simple relations, indicating that the diffusion-based unified model retains nontrivial visual understanding and textual generation capability. While these tasks mostly require lightweight reasoning, given that our model is not trained with large-scale, dedicated reasoning supervision, the results provide a positive signal that the in-image text design can generalize beyond action prediction. These observations suggest clear headroom for scaling \model with more text supervision and longer reasoning, and further highlight the potential of diffusion-based unified models.

\begin{figure*}[t]
    \centering
    \includegraphics[width=1.0\linewidth]{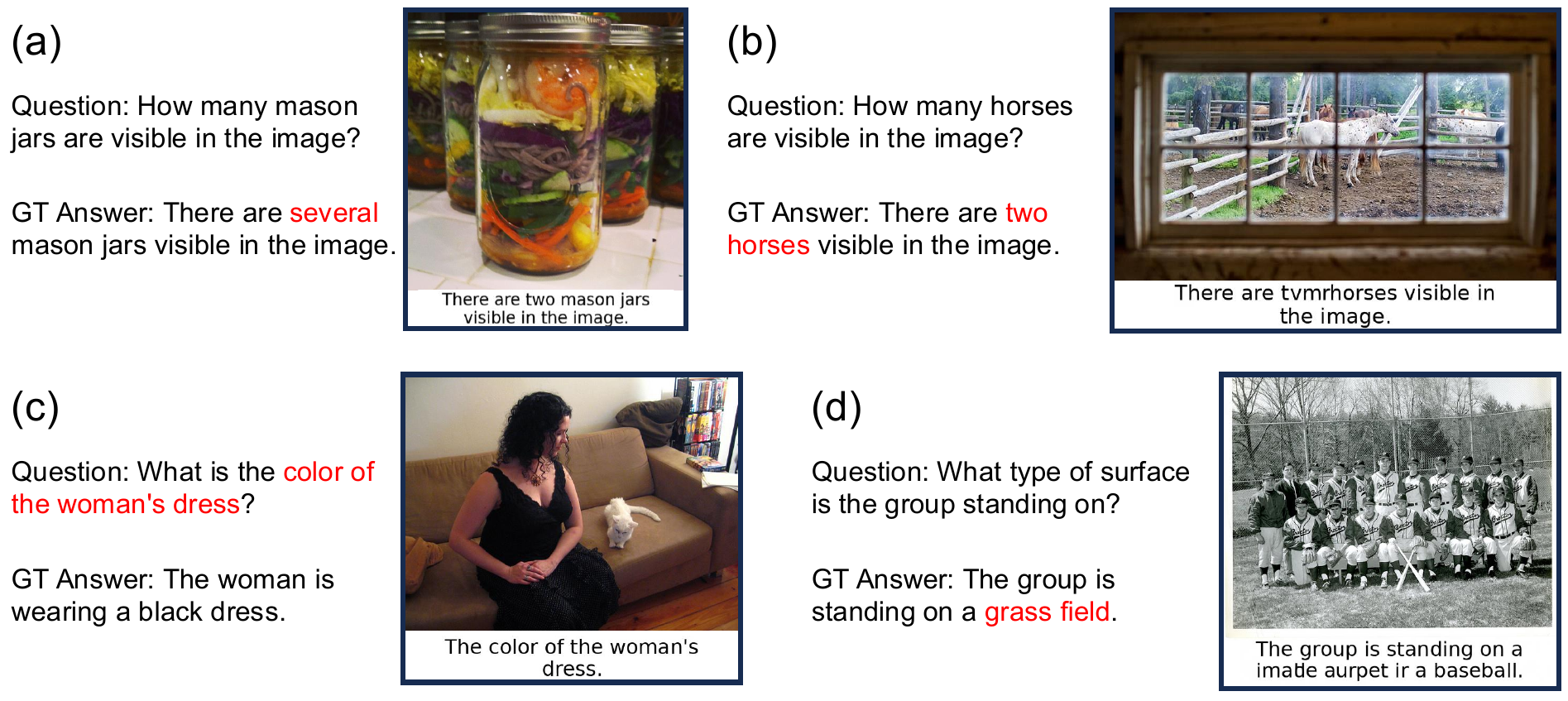}
    \caption{\textbf{Representative failure patterns of text generation.} (a) Counting error. The model produces the wrong number. (b) Text hallucination in in-image rendering. The model captures the intended answer but produces hallucinated characters. (c) Prompt bias in image editing. The model tends to copy salient words from the question instead of generating the inferred answer. (d) Reasoning mismatch. The response is wrong and off-target, but the model still recognizes the scene context, e.g., a baseball team. Error parts are highlighted in \color{red}{red}.}
    \label{fig:text_failure_case}
\end{figure*}

We further provide representative failure cases in Fig.~\ref{fig:text_failure_case}. Fig.~\ref{fig:text_failure_case}-(a) shows a common counting error, where the model predicts an incorrect number despite producing a fluent answer format. Fig.~\ref{fig:text_failure_case}-(b) reflects a text generation failure specific to in-image rendering. The model captures the correct semantic intent but hallucinates or merges characters, resulting in unreadable text, suggesting a positive signal in recognition while leaving clear room for improving typography fidelity. Fig.~\ref{fig:text_failure_case}-(c) illustrates a classic issue in image editing models, where the generation can be influenced by the prompt text itself, leading the model to copy salient words from the question instead of writing the answer it has inferred. Fig.~\ref{fig:text_failure_case}-(d) represents a reasoning mismatch, where the response does not directly answer the asked attribute. Nevertheless, the model correctly identifies the scene as a baseball team on a field, indicating nontrivial visual understanding even when the final verbalization is off-target. Overall, these cases suggest that \model already learns meaningful visual concepts and can often form the right semantic hypothesis, and the remaining errors largely concentrate on faithful text generation and fine-grained instruction following, highlighting the potential of scaling text-in-image supervision and strengthening long-form textual reasoning.

\subsection{Inference Speed}
\label{app:inference-speed}

We report inference time comparisons in Tab.~\ref{exp:inference-speed}. Since \model relies on a diffusion backbone and performs two diffusion passes per step for the two-stage design, it is expected to be slower than autoregressive VLM baselines. Nonetheless, \model is faster than other diffusion-based sequential baselines, benefiting from our lightweight step-wise formulation that conditions only on the most recent frame and avoids carrying the full interaction history. Overall, these results suggest that diffusion-based unified models still have substantial headroom for runtime optimization, and \model provides a practical trade-off between generation quality and inference cost.

\begin{table}[t]
    \centering
    \caption{\textbf{Comparison of Inference Time on VSP.} Runtime is reported as relative inference time per sample on a single H100 GPU, normalized to \texttt{Qwen2.5-VL-7B}. UniCanvas is slower than autoregressive VLMs but faster than other diffusion-based sequential baselines.}
    \scriptsize
    \setlength{\tabcolsep}{4pt}
    \resizebox{0.6\linewidth}{!}{
    \begin{tabular}{@{}lcccc@{}}
        \toprule
        Task & \texttt{Qwen2.5-VL-7B} & MVoT & MindJourney & Ours \\
        \midrule
        VSP & 1.0 & 5.1 & 9.4 & 4.6 \\
        \bottomrule
    \end{tabular}
    }
    \label{exp:inference-speed}
\end{table}

%% file: main.bib
@String(CVPR  = {IEEE Conf. Comput. Vis. Pattern Recog.})

@String(NeurIPS = {Adv. Neural Inform. Process. Syst.})

@String(CVPR  = {CVPR})

@String(NeurIPS = {NeurIPS})

@article{li2025imagine,
  title={Imagine while reasoning in space: Multimodal visualization-of-thought},
  author={Li, Chengzu and Wu, Wenshan and Zhang, Huanyu and Xia, Yan and Mao, Shaoguang and Dong, Li and Vuli{\'c}, Ivan and Wei, Furu},
  journal={arXiv preprint arXiv:2501.07542},
  year={2025}
}

@article{ma2025unitok,
  title={Unitok: A unified tokenizer for visual generation and understanding},
  author={Ma, Chuofan and Jiang, Yi and Wu, Junfeng and Yang, Jihan and Yu, Xin and Yuan, Zehuan and Peng, Bingyue and Qi, Xiaojuan},
  journal={arXiv preprint arXiv:2502.20321},
  year={2025}
}

@article{ma2024janusflow,
  title={Janusflow: Harmonizing autoregression and rectified flow for unified multimodal understanding and generation},
  author={Ma, Yiyang and Liu, Xingchao and Chen, Xiaokang and Liu, Wen and Wu, Chengyue and Wu, Zhiyu and Pan, Zizheng and Xie, Zhenda and Zhang, Haowei and Zhao, Liang and others},
  journal={arXiv preprint arXiv:2411.07975},
  year={2024}
}

@article{team2024chameleon,
  title={Chameleon: Mixed-modal early-fusion foundation models, 2024},
  author={Team, Chameleon},
  journal={URL https://arxiv. org/abs/2405.09818},
  volume={9},
  number={8},
  year={2024}
}

@article{saharia2022photorealistic,
  title={Photorealistic text-to-image diffusion models with deep language understanding},
  author={Saharia, Chitwan and Chan, William and Saxena, Saurabh and Li, Lala and Whang, Jay and Denton, Emily L and Ghasemipour, Kamyar and Gontijo Lopes, Raphael and Karagol Ayan, Burcu and Salimans, Tim and others},
  journal={Advances in neural information processing systems},
  volume={35},
  pages={36479--36494},
  year={2022}
}

@inproceedings{rombach2022high,
  title={High-resolution image synthesis with latent diffusion models},
  author={Rombach, Robin and Blattmann, Andreas and Lorenz, Dominik and Esser, Patrick and Ommer, Bj{\"o}rn},
  booktitle={Proceedings of the IEEE/CVF conference on computer vision and pattern recognition},
  pages={10684--10695},
  year={2022}
}

@article{ramesh2022hierarchical,
  title={Hierarchical text-conditional image generation with clip latents},
  author={Ramesh, Aditya and Dhariwal, Prafulla and Nichol, Alex and Chu, Casey and Chen, Mark},
  journal={arXiv preprint arXiv:2204.06125},
  volume={1},
  number={2},
  pages={3},
  year={2022}
}

@article{podell2023sdxl,
  title={Sdxl: Improving latent diffusion models for high-resolution image synthesis},
  author={Podell, Dustin and English, Zion and Lacey, Kyle and Blattmann, Andreas and Dockhorn, Tim and M{\"u}ller, Jonas and Penna, Joe and Rombach, Robin},
  journal={arXiv preprint arXiv:2307.01952},
  year={2023}
}

@article{gao2024lumina,
  title={Lumina-t2x: Transforming text into any modality, resolution, and duration via flow-based large diffusion transformers},
  author={Gao, Peng and Zhuo, Le and Liu, Dongyang and Du, Ruoyi and Luo, Xu and Qiu, Longtian and Zhang, Yuhang and Lin, Chen and Huang, Rongjie and Geng, Shijie and others},
  journal={arXiv preprint arXiv:2405.05945},
  year={2024}
}

@article{chen2023pixart,
  title={Pixart: Fast training of diffusion transformer for photorealistic text-to-image synthesis},
  author={Chen, Junsong and Yu, Jincheng and Ge, Chongjian and Yao, Lewei and Xie, Enze and Wu, Yue and Wang, Zhongdao and Kwok, James and Luo, Ping and Lu, Huchuan and others},
  journal={arXiv preprint arXiv:2310.00426},
  year={2023}
}

@article{ho2022classifier,
  title={Classifier-free diffusion guidance},
  author={Ho, Jonathan and Salimans, Tim},
  journal={arXiv preprint arXiv:2207.12598},
  year={2022}
}

@article{song2021maximum,
  title={Maximum likelihood training of score-based diffusion models},
  author={Song, Yang and Durkan, Conor and Murray, Iain and Ermon, Stefano},
  journal={Advances in neural information processing systems},
  volume={34},
  pages={1415--1428},
  year={2021}
}

@inproceedings{nichol2021improved,
  title={Improved denoising diffusion probabilistic models},
  author={Nichol, Alexander Quinn and Dhariwal, Prafulla},
  booktitle={International conference on machine learning},
  pages={8162--8171},
  year={2021},
  organization={PMLR}
}

@article{kingma2021variational,
  title={Variational diffusion models},
  author={Kingma, Diederik and Salimans, Tim and Poole, Ben and Ho, Jonathan},
  journal={Advances in neural information processing systems},
  volume={34},
  pages={21696--21707},
  year={2021}
}

@article{karras2022elucidating,
  title={Elucidating the design space of diffusion-based generative models},
  author={Karras, Tero and Aittala, Miika and Aila, Timo and Laine, Samuli},
  journal={Advances in neural information processing systems},
  volume={35},
  pages={26565--26577},
  year={2022}
}

@article{dhariwal2021diffusion,
  title={Diffusion models beat gans on image synthesis},
  author={Dhariwal, Prafulla and Nichol, Alexander},
  journal={Advances in neural information processing systems},
  volume={34},
  pages={8780--8794},
  year={2021}
}

@article{song2020score,
  title={Score-based generative modeling through stochastic differential equations},
  author={Song, Yang and Sohl-Dickstein, Jascha and Kingma, Diederik P and Kumar, Abhishek and Ermon, Stefano and Poole, Ben},
  journal={arXiv preprint arXiv:2011.13456},
  year={2020}
}

@inproceedings{sohl2015deep,
  title={Deep Unsupervised Learning using Nonequilibrium Thermodynamics},
  author={Jascha Sohl{-}Dickstein and Eric A. Weiss and Niru Maheswaranathan and Surya Ganguli},
  booktitle={Proceedings of the 32nd International Conference on Machine Learning},
  pages={2256--2265},
  year={2015},
}

@inproceedings{ho2020ddpm,
  title={Denoising Diffusion Probabilistic Models},
  author={Jonathan Ho and Ajay Jain and Pieter Abbeel},
  booktitle={Advances in Neural Information Processing Systems},
  year={2020},
}

@inproceedings{xie2025showo,
    title={Show-o: One Single Transformer to Unify Multimodal Understanding and Generation},
    author={Jinheng Xie and Weijia Mao and Zechen Bai and David Junhao Zhang and Weihao Wang and Kevin Qinghong Lin and Yuchao Gu and Zhijie Chen and Zhenheng Yang and Mike Zheng Shou},
    booktitle={The Thirteenth International Conference on Learning Representations},
    year={2025}
}

@inproceedings{zhou2025transfusion,
    title={Transfusion: Predict the Next Token and Diffuse Images with One Multi-Modal Model},
    author={Chunting Zhou and Lili Yu and Arun Babu and Kushal Tirumala and Michihiro Yasunaga and Leonid Shamis and Jacob Kahn and Xuezhe Ma and Luke Zettlemoyer and Omer Levy},
    booktitle={The Thirteenth International Conference on Learning Representations},
    year={2025}
}

@article{bai2025qwen25vl,
    title={Qwen2.5-VL Technical Report},
    author={Shuai Bai and Keqin Chen and Xuejing Liu and Jialin Wang and Wenbin Ge and Sibo Song and Kai Dang and Peng Wang and Shijie Wang and others},
    journal={arXiv preprint arXiv:2502.13923},
    year={2025} 
}

@article{dong2023dreamllm,
  title={Dreamllm: Synergistic multimodal comprehension and creation},
  author={Dong, Runpei and Han, Chunrui and Peng, Yuang and Qi, Zekun and Ge, Zheng and Yang, Jinrong and Zhao, Liang and Sun, Jianjian and Zhou, Hongyu and Wei, Haoran and others},
  journal={arXiv preprint arXiv:2309.11499},
  year={2023}
}

@article{chern2024anole,
  title={Anole: An open, autoregressive, native large multimodal models for interleaved image-text generation},
  author={Chern, Ethan and Su, Jiadi and Ma, Yan and Liu, Pengfei},
  journal={arXiv preprint arXiv:2407.06135},
  year={2024}
}

@article{yang2025mmada,
  title={Mmada: Multimodal large diffusion language models},
  author={Yang, Ling and Tian, Ye and Li, Bowen and Zhang, Xinchen and Shen, Ke and Tong, Yunhai and Wang, Mengdi},
  journal={arXiv preprint arXiv:2505.15809},
  year={2025}
}

@article{xu2025visual,
  title={Visual Planning: Let's Think Only with Images},
  author={Xu, Yi and Li, Chengzu and Zhou, Han and Wan, Xingchen and Zhang, Caiqi and Korhonen, Anna and Vuli{\'c}, Ivan},
  journal={arXiv preprint arXiv:2505.11409},
  year={2025}
}

@article{deng2025emerging,
  title={Emerging properties in unified multimodal pretraining},
  author={Deng, Chaorui and Zhu, Deyao and Li, Kunchang and Gou, Chenhui and Li, Feng and Wang, Zeyu and Zhong, Shu and Yu, Weihao and Nie, Xiaonan and Song, Ziang and others},
  journal={arXiv preprint arXiv:2505.14683},
  year={2025}
}

@article{tian2024mm,
  title={Mm-interleaved: Interleaved image-text generative modeling via multi-modal feature synchronizer},
  author={Tian, Changyao and Zhu, Xizhou and Xiong, Yuwen and Wang, Weiyun and Chen, Zhe and Wang, Wenhai and Chen, Yuntao and Lu, Lewei and Lu, Tong and Zhou, Jie and others},
  journal={arXiv preprint arXiv:2401.10208},
  year={2024}
}

@article{wu2024vsp,
  title={Vsp: Assessing the dual challenges of perception and reasoning in spatial planning tasks for vlms},
  author={Wu, Qiucheng and Zhao, Handong and Saxon, Michael and Bui, Trung and Wang, William Yang and Zhang, Yang and Chang, Shiyu},
  journal={arXiv preprint arXiv:2407.01863},
  year={2024}
}

@article{james2020rlbench,
  title={Rlbench: The robot learning benchmark \& learning environment},
  author={James, Stephen and Ma, Zicong and Arrojo, David Rovick and Davison, Andrew J},
  journal={IEEE Robotics and Automation Letters},
  volume={5},
  number={2},
  pages={3019--3026},
  year={2020},
  publisher={IEEE}
}

@article{bai2025qwen2,
  title={Qwen2. 5-vl technical report},
  author={Bai, Shuai and Chen, Keqin and Liu, Xuejing and Wang, Jialin and Ge, Wenbin and Song, Sibo and Dang, Kai and Wang, Peng and Wang, Shijie and Tang, Jun and others},
  journal={arXiv preprint arXiv:2502.13923},
  year={2025}
}

@article{zhen2025tesseract,
    title={TesserAct: Learning 4D Embodied World Models},
    author={Zhen, Haoyu and Sun, Qiao and Zhang, Hongxin and Li, Junyan and Zhou, Siyuan and Du, Yilun and Gan, Chuang},
    journal={arXiv preprint arXiv:2504.20995},
    year={2025}
}

@article{tong2024metamorph,
  title={Metamorph: Multimodal understanding and generation via instruction tuning},
  author={Tong, Shengbang and Fan, David and Zhu, Jiachen and Xiong, Yunyang and Chen, Xinlei and Sinha, Koustuv and Rabbat, Michael and LeCun, Yann and Xie, Saining and Liu, Zhuang},
  journal={arXiv preprint arXiv:2412.14164},
  year={2024}
}

@article{chen2025blip3,
  title={Blip3-o: A family of fully open unified multimodal models-architecture, training and dataset},
  author={Chen, Jiuhai and Xu, Zhiyang and Pan, Xichen and Hu, Yushi and Qin, Can and Goldstein, Tom and Huang, Lifu and Zhou, Tianyi and Xie, Saining and Savarese, Silvio and others},
  journal={arXiv preprint arXiv:2505.09568},
  year={2025}
}

@article{wang2025fudoki,
  title={FUDOKI: Discrete Flow-based Unified Understanding and Generation via Kinetic-Optimal Velocities},
  author={Wang, Jin and Lai, Yao and Li, Aoxue and Zhang, Shifeng and Sun, Jiacheng and Kang, Ning and Wu, Chengyue and Li, Zhenguo and Luo, Ping},
  journal={arXiv:2505.20147},
  year={2025}
}

@article{shi2025muddit,
  title={Muddit: Liberating Generation Beyond Text-to-Image with a Unified Discrete Diffusion Model},
  author={Shi, Qingyu and Bai, Jinbin and Zhao, Zhuoran and Chai, Wenhao and Yu, Kaidong and Wu, Jianzong and Song, Shuangyong and Tong, Yunhai and Li, Xiangtai and Li, Xuelong and others},
  journal={arXiv:2505.23606},
  year={2025}
}

@article{duan2025got,
  title={GoT-R1: Unleashing Reasoning Capability of MLLM for Visual Generation with Reinforcement Learning},
  author={Duan, Chengqi and Fang, Rongyao and Wang, Yuqing and Wang, Kun and Huang, Linjiang and Zeng, Xingyu and Li, Hongsheng and Liu, Xihui},
  journal={arXiv preprint arXiv:2505.17022},
  year={2025}
}

@article{jiang2025t2i,
  title={T2i-r1: Reinforcing image generation with collaborative semantic-level and token-level cot},
  author={Jiang, Dongzhi and Guo, Ziyu and Zhang, Renrui and Zong, Zhuofan and Li, Hao and Zhuo, Le and Yan, Shilin and Heng, Pheng-Ann and Li, Hongsheng},
  journal={arXiv preprint arXiv:2505.00703},
  year={2025}
}

@article{zhang2026foreact,
  title={ForeAct: Steering Your VLA with Efficient Visual Foresight Planning},
  author={Zhang, Zhuoyang and Yang, Shang and Hu, Qinghao and Huang, Luke J and Hou, James and Sun, Yufei and Lu, Yao and Han, Song},
  journal={arXiv preprint arXiv:2602.12322},
  year={2026}
}

@article{jiang2025co,
  title={Co-reinforcement learning for unified multimodal understanding and generation},
  author={Jiang, Jingjing and Si, Chongjie and Luo, Jun and Zhang, Hanwang and Ma, Chao},
  journal={arXiv preprint arXiv:2505.17534},
  year={2025}
}

@article{qin2025unicot,
  title={Uni-cot: Towards unified chain-of-thought reasoning across text and vision},
  author={Qin, Luozheng and Gong, Jia and Sun, Yuqing and Li, Tianjiao and Yang, Mengping and Yang, Xiaomeng and Qu, Chao and Tan, Zhiyu and Li, Hao},
  journal={arXiv preprint arXiv:2508.05606},
  year={2025}
}

@article{zhen20243d,
  title={3d-vla: A 3d vision-language-action generative world model},
  author={Zhen, Haoyu and Qiu, Xiaowen and Chen, Peihao and Yang, Jincheng and Yan, Xin and Du, Yilun and Hong, Yining and Gan, Chuang},
  journal={arXiv preprint arXiv:2403.09631},
  year={2024}
}

@article{gao2025adaworld,
  title={Adaworld: Learning adaptable world models with latent actions},
  author={Gao, Shenyuan and Zhou, Siyuan and Du, Yilun and Zhang, Jun and Gan, Chuang},
  journal={arXiv preprint arXiv:2503.18938},
  year={2025}
}

@article{liao2025genie,
  title={Genie envisioner: A unified world foundation platform for robotic manipulation},
  author={Liao, Yue and Zhou, Pengfei and Huang, Siyuan and Yang, Donglin and Chen, Shengcong and Jiang, Yuxin and Hu, Yue and Cai, Jingbin and Liu, Si and Luo, Jianlan and others},
  journal={arXiv preprint arXiv:2508.05635},
  year={2025}
}

@inproceedings{bar2025navigation,
  title={Navigation world models},
  author={Bar, Amir and Zhou, Gaoyue and Tran, Danny and Darrell, Trevor and LeCun, Yann},
  booktitle={Proceedings of the Computer Vision and Pattern Recognition Conference},
  pages={15791--15801},
  year={2025}
}

@article{yang2025mindjourney,
  title={MindJourney: Test-Time Scaling with World Models for Spatial Reasoning},
  author={Yang, Yuncong and Liu, Jiageng and Zhang, Zheyuan and Zhou, Siyuan and Tan, Reuben and Yang, Jianwei and Du, Yilun and Gan, Chuang},
  journal={arXiv preprint arXiv:2507.12508},
  year={2025}
}

@article{qian2025wristworld,
  title={Wristworld: Generating wrist-views via 4d world models for robotic manipulation},
  author={Qian, Zezhong and Chi, Xiaowei and Li, Yuming and Wang, Shizun and Qin, Zhiyuan and Ju, Xiaozhu and Han, Sirui and Zhang, Shanghang},
  journal={arXiv preprint arXiv:2510.07313},
  year={2025}
}

@misc{ye2026worldactionmodelszeroshot,
      title={World Action Models are Zero-shot Policies},
      author={Seonghyeon Ye and Yunhao Ge and Kaiyuan Zheng and Shenyuan Gao and Sihyun Yu and George Kurian and Suneel Indupuru and You Liang Tan and Chuning Zhu and Jiannan Xiang and Ayaan Malik and Kyungmin Lee and William Liang and Nadun Ranawaka and Jiasheng Gu and Yinzhen Xu and Guanzhi Wang and Fengyuan Hu and Avnish Narayan and Johan Bjorck and Jing Wang and Gwanghyun Kim and Dantong Niu and Ruijie Zheng and Yuqi Xie and Jimmy Wu and Qi Wang and Ryan Julian and Danfei Xu and Yilun Du and Yevgen Chebotar and Scott Reed and Jan Kautz and Yuke Zhu and Linxi "Jim" Fan and Joel Jang},
      year={2026},
      eprint={2602.15922},
      archivePrefix={arXiv},
      primaryClass={cs.RO},
      url={https://arxiv.org/abs/2602.15922},
}

@inproceedings{zeng2025editworld,
  title={Editworld: Simulating world dynamics for instruction-following image editing},
  author={Zeng, Bohan and Yang, Ling and Liu, Jiaming and Xu, Minghao and Zhang, Yuanxing and Wan, Pengfei and Zhang, Wentao and Yan, Shuicheng},
  booktitle={Proceedings of the 33rd ACM International Conference on Multimedia},
  pages={12674--12681},
  year={2025}
}

@article{teoh2025next,
  title={Next-Latent Prediction Transformers Learn Compact World Models},
  author={Teoh, Jayden and Tomar, Manan and Ahn, Kwangjun and Hu, Edward S and Sharma, Pratyusha and Islam, Riashat and Lamb, Alex and Langford, John},
  journal={arXiv preprint arXiv:2511.05963},
  year={2025}
}

@article{lingbot-va2026,
  title={Causal World Modeling for Robot Control},
  author={Li, Lin and Zhang, Qihang and Luo, Yiming and Yang, Shuai and Wang, Ruilin and Han, Fei and Yu, Mingrui and Gao, Zelin and Xue, Nan and Zhu, Xing and Shen, Yujun and Xu, Yinghao},
  journal={arXiv preprint arXiv:2601.21998},
  year={2026}
}

@article{hafner2019dream,
  title={Dream to control: Learning behaviors by latent imagination},
  author={Hafner, Danijar and Lillicrap, Timothy and Ba, Jimmy and Norouzi, Mohammad},
  journal={arXiv preprint arXiv:1912.01603},
  year={2019}
}

@article{ha2018world,
  title={World models},
  author={Ha, David and Schmidhuber, J{\"u}rgen},
  journal={arXiv preprint arXiv:1803.10122},
  volume={2},
  number={3},
  pages={440},
  year={2018}
}

@article{gu2025thinkmorph,
  title={ThinkMorph: Emergent Properties in Multimodal Interleaved Chain-of-Thought Reasoning},
  author={Gu, Jiawei and Hao, Yunzhuo and Wang, Huichen Will and Li, Linjie and Shieh, Michael Qizhe and Choi, Yejin and Krishna, Ranjay and Cheng, Yu},
  journal={arXiv preprint arXiv:2510.27492},
  year={2025}
}

@article{wu2026visual,
  title={Visual Generation Unlocks Human-Like Reasoning through Multimodal World Models},
  author={Wu, Jialong and Zhang, Xiaoying and Yuan, Hongyi and Zhang, Xiangcheng and Huang, Tianhao and He, Changjing and Deng, Chaoyi and Zhang, Renrui and Wu, Youbin and Long, Mingsheng},
  journal={arXiv preprint arXiv:2601.19834},
  year={2026}
}

@article{xiao2025mindomni,
  title={Mindomni: Unleashing reasoning generation in vision language models with rgpo},
  author={Xiao, Yicheng and Song, Lin and Chen, Yukang and Luo, Yingmin and Chen, Yuxin and Gan, Yukang and Huang, Wei and Li, Xiu and Qi, Xiaojuan and Shan, Ying},
  journal={arXiv preprint arXiv:2505.13031},
  year={2025}
}

@article{wu2024vila,
  title={Vila-u: a unified foundation model integrating visual understanding and generation},
  author={Wu, Yecheng and Zhang, Zhuoyang and Chen, Junyu and Tang, Haotian and Li, Dacheng and Fang, Yunhao and Zhu, Ligeng and Xie, Enze and Yin, Hongxu and Yi, Li and others},
  journal={arXiv preprint arXiv:2409.04429},
  year={2024}
}

@inproceedings{esser2024scaling,
  title={Scaling rectified flow transformers for high-resolution image synthesis},
  author={Esser, Patrick and Kulal, Sumith and Blattmann, Andreas and Entezari, Rahim and M{\"u}ller, Jonas and Saini, Harry and Levi, Yam and Lorenz, Dominik and Sauer, Axel and Boesel, Frederic and others},
  booktitle={Forty-first international conference on machine learning},
  year={2024}
}

@article{yang2025machine,
  title={Machine Mental Imagery: Empower Multimodal Reasoning with Latent Visual Tokens},
  author={Yang, Zeyuan and Yu, Xueyang and Chen, Delin and Shen, Maohao and Gan, Chuang},
  journal={arXiv preprint arXiv:2506.17218},
  year={2025}
}

@inproceedings{zhang2018perceptual,
  title={The Unreasonable Effectiveness of Deep Features as a Perceptual Metric},
  author={Zhang, Richard and Isola, Phillip and Efros, Alexei A and Shechtman, Eli and Wang, Oliver},
  booktitle={CVPR},
  year={2018}
}

@inproceedings{liu2023llava,
    author      = {Liu, Haotian and Li, Chunyuan and Wu, Qingyang and Lee, Yong Jae},
    title       = {Visual Instruction Tuning},
    booktitle   = {NeurIPS},
    year        = {2023}
  }

@article{wei2024omniedit,
  title={OmniEdit: Building Image Editing Generalist Models Through Specialist Supervision},
  author={Wei, Cong and Xiong, Zheyang and Ren, Weiming and Du, Xinrun and Zhang, Ge and Chen, Wenhu},
  journal={arXiv preprint arXiv:2411.07199},
  year={2024}
}

@article{guo2025thinking,
  title={Thinking-while-generating: Interleaving textual reasoning throughout visual generation},
  author={Guo, Ziyu and Zhang, Renrui and Li, Hongyu and Zhang, Manyuan and Chen, Xinyan and Wang, Sifan and Feng, Yan and Pei, Peng and Heng, Pheng-Ann},
  journal={arXiv preprint arXiv:2511.16671},
  year={2025}
}

@article{guo2025can,
  title={Can We Generate Images with CoT? Let's Verify and Reinforce Image Generation Step by Step},
  author={Guo, Ziyu and Zhang, Renrui and Tong, Chengzhuo and Zhao, Zhizheng and Huang, Rui and Zhang, Haoquan and Zhang, Manyuan and Liu, Jiaming and Zhang, Shanghang and Gao, Peng and others},
  journal={arXiv preprint arXiv:2501.13926},
  year={2025}
}

@article{jiang2025draco,
  title={DraCo: Draft as CoT for Text-to-Image Preview and Rare Concept Generation},
  author={Jiang, Dongzhi and Zhang, Renrui and Li, Haodong and Zong, Zhuofan and Guo, Ziyu and He, Jun and Guo, Claire and Ye, Junyan and Fang, Rongyao and Li, Weijia and others},
  journal={arXiv preprint arXiv:2512.05112},
  year={2025}
}

@article{han2026unicorn,
  title={UniCorn: Towards Self-Improving Unified Multimodal Models through Self-Generated Supervision},
  author={Han, Ruiyan and Fang, Zhen and Sun, XinYu and Ma, Yuchen and Wang, Ziheng and Zeng, Yu and Chen, Zehui and Chen, Lin and Huang, Wenxuan and Xu, Wei-Jie and others},
  journal={arXiv preprint arXiv:2601.03193},
  year={2026}
}

@article{chen2026show,
  title={Show, Don't Tell: Morphing Latent Reasoning into Image Generation},
  author={Chen, Harold Haodong and Yin, Xinxiang and Shu, Wen-Jie and Zhang, Hongfei and Zhang, Zixin and Liao, Chenfei and Guo, Litao and Chen, Qifeng and Chen, Ying-Cong},
  journal={arXiv preprint arXiv:2602.02227},
  year={2026}
}

@article{su2026generation,
  title={Generation Enhances Understanding in Unified Multimodal Models via Multi-Representation Generation},
  author={Su, Zihan and Wei, Hongyang and Cen, Kangrui and Wang, Yong and Chen, Guanhua and Yuan, Chun and Chu, Xiangxiang},
  journal={arXiv preprint arXiv:2601.21406},
  year={2026}
}

@article{yin2025reasonedit,
  title={ReasonEdit: Towards Reasoning-Enhanced Image Editing Models},
  author={Yin, Fukun and Liu, Shiyu and Han, Yucheng and Wang, Zhibo and Xing, Peng and Wang, Rui and Cheng, Wei and Wang, Yingming and Li, Aojie and Yin, Zixin and others},
  journal={arXiv preprint arXiv:2511.22625},
  year={2025}
}

@article{wang2026promptrl,
  title={PromptRL: Prompt Matters in RL for Flow-Based Image Generation},
  author={Wang, Fu-Yun and Zhang, Han and Gharbi, Michael and Li, Hongsheng and Park, Taesung},
  journal={arXiv preprint arXiv:2602.01382},
  year={2026}
}
